\documentclass[pmlr]{jmlr}
\usepackage{mathtools}
\usepackage{multirow}
\usepackage{enumitem}

\usepackage{longtable}

\usepackage{booktabs}
\usepackage[load-configurations=version-1]{siunitx} 

\makeatletter
\def\set@curr@file#1{\def\@curr@file{#1}} 
\makeatother

\theorembodyfont{\upshape}
\theoremheaderfont{\scshape}
\theorempostheader{:}
\theoremsep{\newline}

\usepackage{algorithmic,algorithm}

\title[Skillearn: Machine Learning Inspired by Humans' Learning Skills]{Skillearn: Machine Learning Inspired by Humans' Learning Skills}

\author{\Name{Pengtao Xie}
       \Email{p1xie@eng.ucsd.edu} 
       \AND
\Name{Xuefeng Du}
       \Email{xuefengdu1@gmail.com} 
\AND
\Name{Hao Ban}
       \Email{bhsimon0810@gmail.com} 
       \AND
       \addr University of California San Diego
       }

\begin{document}

\maketitle

\begin{abstract}
Humans, as the most powerful learners on the planet, have accumulated a lot of learning skills, such as learning through tests, interleaving learning,  self-explanation, active recalling, to name a few. These learning skills and methodologies enable humans to learn new topics more effectively and efficiently. We are interested in investigating whether humans' learning skills can be borrowed to help machines to learn better. Specifically, we aim to formalize these skills and leverage them to train better machine learning (ML) models. To achieve this goal, we develop a general framework -- Skillearn, which provides a principled way to represent humans' learning skills mathematically and use the formally-represented skills to improve the training of ML models. 
In two case studies, we apply Skillearn to formalize two learning skills of humans: learning by passing tests and interleaving learning, and use the formalized skills to improve neural architecture search. Experiments on various datasets show that trained using the skills formalized by Skillearn, ML models achieve significantly better performance. 
\end{abstract}

\section{Introduction}
Given a group of human students, assuming they work equally hard, there are three major factors determining which students learn better than others, including intelligence, learning skills, and learning materials. People with higher intelligence quotient (IQ) are stronger learners. Learning materials, such as textbooks, video lectures, practice questions, etc. are also crucial in determining the quality of learning. Another vital factor impacting learning outcomes is learning skills. Oftentimes, students in the same class have similar IQ and have access to the same learning materials, but their final grades (which measure learning quality) have a large variance. The major differentiating factor is that different students have different levels of mastery of learning skills. Some students have better learning methodologies, which enable them to learn faster and better. In the long history of learning, humans have accumulated a lot of effective learning skills, such as learning through tests, interleaving learning, self-explanation, active recalling, etc. 

\begin{figure*}[t]
    \centering
 \includegraphics[width=0.9\textwidth]{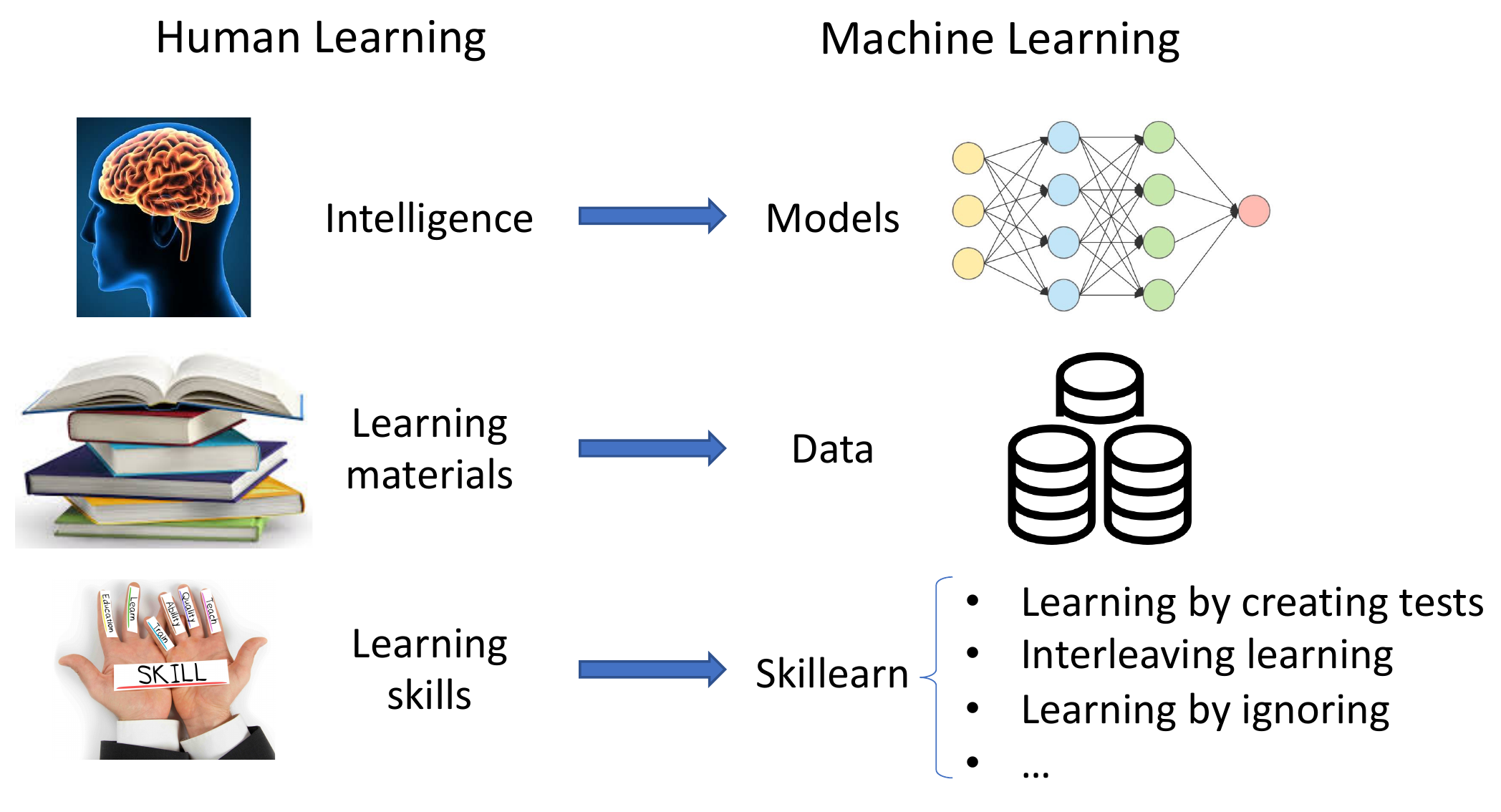}
       \caption{Human learning (HL) versus machine learning (ML). Model capacity in ML is analogous to intelligence in HL. Data in ML is analogous to learning materials in HL. The machines' learning skills formulated by our proposed Skillearn framework are analogous to humans' learning skills. 
}
 \label{fig:hl-ml}
\end{figure*}

Similar to human learning, the performance of machine learning (ML) models is also determined by several factors. In the current practice of ML, two dominant factors determining ML performance are the capacity of models and the abundance of data. 
ML model capacity is analogous to the intelligence of humans. From linear models such as support vector machine to nonlinear models such as deep neural networks, ML researchers have been continuously building more powerful ML models to deal with more complicated tasks. It is like the evolution of humans' brains, which become increasingly intelligent. Data for ML is analogous to learning materials for humans. ML models trained with more labeled data in general perform better.

For intelligence and learning materials in human learning (HL), we identify their counterparts in machine learning as model capacity and data. We are interested in asking: for learning skills in HL, do they have counterparts in ML as well? Can machines be equipped with effective learning skills as humans are? In this paper, we aim to address these questions. We propose a general framework -- Skillearn, which draws inspiration from humans' learning skills and formulates them into machines' learning skills (MLS). These MLS are leveraged to train better ML models. In Skillearn, there are one or multiple learner models, each with one or multiple sets of learnable parameters such as weight parameters, architectures, hyperparameters, etc. Different learners interact with each other through interaction functions. The learning of all learners is organized into multiple stages, each involving a subset of learners. The stages have an order, but they are  performed end-to-end in a multi-level optimization framework where latter stages influence earlier stages and vice versa. We develop a unified optimization algorithm for solving the multi-level optimization problem in Skillearn. In two case studies, we apply Skillearn to formalize two learning skills of humans -- learning by passing tests (LPT) and interleaving learning (IL) -- into machines' learning skills (MLS) and leverage these MLS for neural architecture search~\citep{zoph2016neural,real2019regularized,liu2018darts}. In LPT, a tester model dynamically creates tests with increasing levels of difficulty to evaluate a testee model; the testee continuously improves its architecture by passing however difficult tests created by the tester. In IL, a set of  models collaboratively learn a data encoder in an interleaving fashion: the encoder is trained by model 1 for a while, then passed to model 2 for further training, then model 3, and so on; after trained by all models, the encoder returns back to model 1 and is trained again, then moving to model 2, 3, etc. This process repeats for multiple rounds. Experiments on various datasets demonstrate that ML models trained by these two learning skills achieve significantly better performance.

The major contributions of this work are as follows.

\begin{itemize}
    \item We propose to leverage the broadly-used and effective learning skills in human learning to develop better machine learning methods.
    \item We propose Skillearn, a general framework for formulating humans' learning skills into machines' learning skills that can be leveraged by ML models for achieving better learning outcomes. 
    \item We apply Skillearn to formalize two skills in human learning -- learning by passing tests (LPT) and interleaving learning (IL), and apply them to improve neural architecture search. 
    \item On various datasets, we demonstrate the effectiveness of the two skills -- LPT and IL formalized by Skillearn -- in learning better neural architectures. 
\end{itemize}

The rest of the paper is organized as follows. Section 2 presents the general Skillearn framework. In Section 3 and 4, we present two case studies, where Skillearn is applied to formalize two skills in human learning: learning by passing tests and interleaving learning. Section 5 reviews related works and Section 6 concludes the paper.

\section{Skillearn: Machine Learning Inspired by Human's Learning Skills}
In this section, we present a general framework called Skillearn, which gets inspiration from humans' learning skills, formalize these skills, and leverage them to improve machine learning. 
We begin with a brief overview of humans' learning skills and summarize their properties. Then we present the Skillearn framework and  the optimization algorithm for this framework. 

\subsection{Humans' Learning Skills}

Humans, as the most powerful learners on the planet, have accumulated a lot of skills and techniques in learning faster and better. Here are some examples. 
\begin{itemize}
\item \textbf{Learning through testing.} After learning a topic, a student can solve some test problems (created or selected by a teacher) about this topic to identify the strong and weak points in his/her understanding of this topic, and re-learn the topic based on the identified strong and weak points. In re-learning, the identified strong and weak points help the student to know what to focus on. The quality of test problems plays a crucial role in effectively evaluating the student. How to create or select high-quality test problems is an important skill that the teacher needs to learn. 
\item \textbf{Interleaving learning} is a learning technique where a learner interleaves the studies of multiple topics: study topic $A$ for a while, then switch to $B$, subsequently to $C$; then switch back to $A$, and so on, forming a pattern of $ABCABCABC\cdots$. Interleaving learning is in contrast to blocked learning, which studies one topic very thoroughly before moving to another topic. Compared with blocked learning, interleaving learning increases long-term retention and improves ability to transfer learned knowledge. 
    \item \textbf{Learning by ignoring.} In course learning, given a large collection of practice problems provided in the textbook, the teacher selects a subset of problems as homework for the students to practice instead of using all problems in the textbook. Some practice problems are ignored because 1) they are too difficult which might confuse the students; 2) they are too simple which are not effective in helping the students to practice their knowledge learned during lectures; 3) they are repetitive. 
\end{itemize}

\subsubsection{Properties of Humans'  Learning Skills}
From the above examples of humans'  learning skills, we observe the following properties of them. 
\begin{itemize}
    \item A learning event involves multiple learners.  
    For example, in learning through testing, there are two learners: a student and a teacher. The teacher learns how to create test problems and the student learns how to solve these test problems. 
    \item In a learning task, a learner has multiple aspects to learn about this task. For example, in learning by ignoring, to create effective homework problems, the teacher needs to learn: 1) how to solve these problems; 2) which problems are more valuable to use as homework. 
      \item Different learners interact with each other during learning. For example, in learning through testing, the teacher creates test problems and uses them to evaluate the student. 
    \item In a learning task, the learning process is divided into multiple stages. These stages have a certain order. Each stage involves a subset of learners. For example, in learning through testing, there are three stages: 1) the teacher learns a topic; 2) the teacher creates test problems about this topic and uses them to evaluate the student; 3) based on the strong and weak points identified during solving the test problems, the student re-learns this topic. 
    The three stages have a sequential order and cannot be switched. The first stage involves the teacher only; the second stage involves both the teacher and the student; the third stage involves the student only. 
    \item Testing and validation are  widely used to evaluate the outcome of learning and provide feedback for improving learning. For example, in learning through testing, the student takes a test to identify the strong and weak points in his/her learning of a topic. 
        \item Learning is performed on various learning materials, including textbooks used for initial learning, homeworks used for enhancing the understanding of knowledge learned from textbooks, tests used for evaluating the outcome of learning, etc. 
\end{itemize}

\subsection{General Framework of Skillearn}

\begin{figure}[H]
    \centering
 \includegraphics[width=\textwidth]{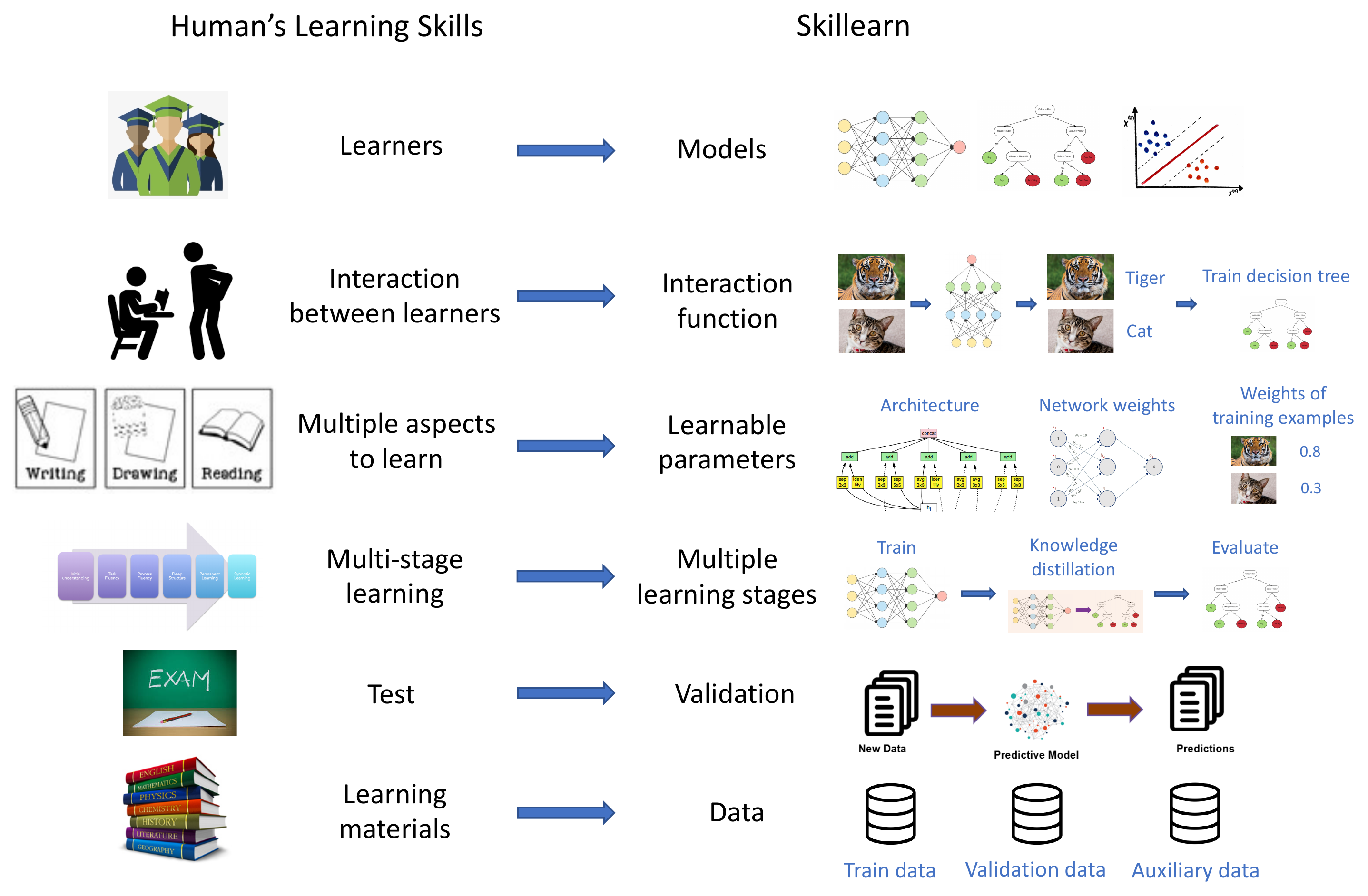}
       \caption{\scriptsize The elements of Skillearn and their counterparts in human learning. The goal of Skillearn is to learn one or a set of ML models which are  analogous to learners in human learning. The models can be of any type, such as deep neural network, decisions tree, support vector machine, etc. In a human learning event, a learner has multiple aspects to learn, such as how to read, how to write, how to draw, etc. Analogously, a model in Skillearn has multiple sets of parameters that are learnable, such as architectures, network weights, weights of training examples, etc. In human learning, different learners interact with each other. For example, a teacher can teach a student. An examiner can evaluate an examinee. Likewise, the models in Skillearn can interact with each other. For example, in knowledge distillation, a teacher model (e.g., a deep neural network) can ``teach" a student model (e.g., a decision tree) where the teacher predicts pseudo labels on unlabeled data, then these pseudo-labeled data examples are used to train the student model. In a human learning event, there are multiple stages of learning  events. For example, in classroom learning, there could be three learning stages: 1) a teacher learns the course materials; 2) the teacher teaches these materials to students; 3) the students take tests to evaluate how well they learn. Analogously, in Skillearn, the learning involves multiple stages. For example, in knowledge distillation, there could be three learning stages: 1) a teacher model is trained; 2) the teacher performs knowledge distillation to ``teach" a student model as described above; 3) the performance of the student model is evaluated. In human learning, tests are widely used to evaluate the learners and provide feedback for improving the learners. Analogously, ML models are validated for further improvement. In human learning, the learners learn from learning materials such as textbooks, lecture notes, homework, etc. Likewise, ML models are learned on various datasets, such as training data, validation data, and other auxiliary data.
       }
 \label{fig:hl-ml}
\end{figure}

Based on the properties of humans'  learning skills, we propose a framework called Skillearn to formalize the learning skills of humans and incorporate them into machine learning.  In Skillearn, we have the following elements. 
\begin{itemize}
    \item \textbf{Learners}. There could be one or multiple learners. Each learner is an ML model, such as a deep convolutional network, a deep generative model, a nonparametric kernel density estimator, etc. 
    This is analogous to human learning which involves one or multiple human learners. 
    \item \textbf{Learnable parameters}. Each learner has one or more sets of learnable parameters, which could be weight parameters of a network, architecture of a network, weights of training examples, hyperparameters, etc. This is analogous to human learning where each human learner learns multiple aspects in a learning task. 
    \item \textbf{Interaction function}, which describes how two or more learners  interact with others. Some examples of interaction include: 1) in knowledge distillation, given an unlabeled image dataset, model $A$ predicts the pseudo labels of these images; then model $B$ is trained using these images and the pseudo labels generated by model $A$; 2) given a set of texts, two text encoders $A$ and $B$ extract embeddings of the texts; $A$ and $B$ are tied together via distributional matching: the distribution of embeddings extracted by $A$ is encouraged to have small total-variance with the distribution of embeddings extracted by $B$. 
    This is analogous to human learning where multiple human learners interact with each other. 
    \item \textbf{Learning stages}. The learning of all learners is not conducted at one shot simultaneously. The learning is performed at multiple stages with an order. 
    At each stage, a subset of learners participate in the learning. For example, in knowledge distillation, there are two stages: 1) a teacher model is trained; 2) the teacher model predicts pseudo labels on an unlabeled dataset and the pseudo-labeled dataset is used to train the student model. The first stage involves a single learner, which is the teacher. The second stage involves two learners: the teacher and the student.  
    This is analogous to human learning where the learning process is divided into multiple stages. Mathematically, we formulate the learning at each stage as an optimization problem. The outcome of one learning stage is passed to another learning stage via the interaction function. 
    \item \textbf{Validation stage.} This stage evaluates the outcome of learning and provides feedback to improve the learning at the learning stages. This is analogous to the testing and validation in human learning. The validation stage is formulated as an optimization problem as well. The learning outcomes produced in the learning stages are passed to the validation stage. 
    \item \textbf{Datasets.} Datasets in ML are analogous to learning materials in human learning. Each learner has a training dataset and a validation dataset. The training dataset is used in the learning stages and the validation dataset is used in the validation stage. Besides, there are auxiliary datasets (labeled or unlabeled) on which the learners interact with each other. 
\end{itemize}

Next, we define the learning stages. Each learning stage performs a focused learning activity which is defined as an optimization problem. The optimization problem involves a training loss and (optionally) an interaction function which describes how the learners involved in this stage interact with each other. A learning stage consists of the following elements:
\begin{itemize}
    \item \textbf{Active learners}. 
    A subset of learners (one or more) are involved at this learning stage. These learners are called active learners.
    \item \textbf{Active learnable parameters}. For each active learner, a sub-collection of its learnable parameter sets are trained in this stage.
    \item \textbf{Supporting learnable parameters}. For each active learner, a sub-collection of its learnable parameter sets are used to define the loss function and interaction function, but they are not updated at this stage. 
    \item \textbf{Active training datasets}, which include the training dataset of every active learner.
    \item \textbf{Active auxiliary datasets}, which include the auxiliary datasets where the interaction function in this learning stage is defined on. 
    \item \textbf{Training loss}, which is defined on the active training data collection, active learnable parameters, and supporting learnable parameters. 
    \item \textbf{Interaction function}, which depicts the interaction between two or more active learners. It is defined on the active auxiliary datasets, active learnable parameters, and supporting learnable parameters.
\end{itemize}

In Skillearn, there is a single validation stage where an optimization problem is defined. The optimization problem involves one or more validation losses and (optionally) an interaction function which describes how the learners in the validation stage interact with each other. The validation stage consists of the following elements. 
\begin{itemize}
\item \textbf{Active learners}, which are the learners to validate. 
    \item  \textbf{Remaining learnable parameters}. At each learning stage, a subset of parameters are learned. After all learning stages, the parameters that have not been learned are called remaining parameters. The remaining parameters are updated in the validation stage.
    \item \textbf{Validation datasets}: validation datasets of all active  learners.
    \item \textbf{Active auxiliary datasets}, which include the auxiliary datasets where the interaction function in the validation stage is defined on. 
    \item \textbf{Validation losses}, which are defined on remaining learnable  parameters, validation datasets,  and (optionally) active  auxiliary datasets.
    \item \textbf{Interaction function}, which depicts the interaction between two or more active learners. It is defined on remaining learnable  parameters and active auxiliary datasets. 
\end{itemize}

\subsubsection{Mathematical Setup}

\begin{table}[t]
\centering
\begin{tabular}{l|p{12cm}}
\hline
Notation & Meaning \\
\hline
$M$ & Number of learners\\
$D_m^{(\textrm{tr})}$ & Training data of the $m$-th learner\\
$D_m^{(\textrm{val})}$ & Validation data of the $m$-th learner\\
$\mathcal{F}$ & Auxiliary datasets accessible to all learners\\
$N_m$ & Number of learnable parameter sets belonging to the $m$-th learner\\
$W_i^{(m)}$ & The $i$-th learnable parameter set of the $m$-th learner\\
$K$ & Number of learning
stages\\
$M_k$ & Number of active learners in the $k$-th learning stage\\
$\mathcal{A}_k$ & The set of active learners in the $k$-th learning stage\\
$a_i^{(k)}$ & The $i$-th active learner in the $k$-th learning stage\\
$O_{ki}$ & The number of active parameter sets of the $i$-th active learner in the $k$-th learning stage\\
$W_{kij}$ & The $j$-th active parameter set of the $i$-th learner in the $k$-th learning stage \\
$\mathcal{W}_{ki}$ & The collection of active parameter sets of the $i$-th active learner in the $k$-th learning stage \\
$\mathcal{W}_k$ & All active parameter sets in the $k$-th learning stage \\
$P_{ki}$ & The number of supporting parameter sets of the $i$-th active learner in the $k$-th learning stage\\
$U_{kij}$ & The $j$-th supporting parameter set of the $i$-th active  learner in the $k$-th learning stage \\
$\mathcal{U}_{ki}$ & The collection of supporting parameter sets of the $i$-th active learner in the $k$-th learning stage \\
$\mathcal{U}_k$ & All supporting parameter sets in the $k$-th learning stage \\
$D_{ki}^{(\textrm{tr})}$ & Training dataset of the $i$-th active learner in the $k$-th learning stage\\
$\mathcal{D}_k^{(\textrm{tr})}$ & Active training datasets in the $k$-th learning stage\\
$\mathcal{F}_k$ & Active auxiliary datasets in the $k$-th learning stage\\
$L_k$ & Training loss in the $k$-th learning stage\\
$I_k$ & Interaction function in the $k$-th learning stage\\
\hline
\end{tabular}
\caption{Notations in the Skillearn framework}
\label{tb:skl_notations}
\end{table}

We assume there are $M$ learners in total. For each learner $m$, it has a training set $D^{(\textrm{tr})}_m$ and optionally a validation set $D^{(\textrm{val})}_m$. Meanwhile, all learners  share a common collection of auxiliary datasets $\mathcal{F}$, which could be unlabeled datasets used for self-supervised pretraining~\citep{he2019moco}, additional labeled datasets used for validation, and so on. The learner  $m$ has one or more sets of learnable parameters $\{W_i^{(m)}\}_{i=1}^{N_m}$. The learnable parameters could be network weights, architectures, hyperparameters, weights of training examples, etc. 

We assume there are $K$ learning stages. At each stage $k$, a subset of $M_k$ learners  $\mathcal{A}_k=\{a^{(k)}_i\}_{i=1}^{M_k}$ are involved in the learning, which are called active learners. For each active learner $a^{(k)}_i$, a sub-collection of its learnable parameter sets $\mathcal{W}_{ki}=\{W_{kij}\}_{j=1}^{O_{ki}}$ are trained at this stage, which are called active learnable parameters. Let $\mathcal{W}_{k}=\{\mathcal{W}_{ki}|i=1,\cdots,M_k\}$ denote the active learnable parameters for all active learners. Meanwhile, another sub-collection of its learnable parameter sets $\mathcal{U}_{ki}=\{U_{kij}\}_{j=1}^{P_{ki}}$ are used to define the training loss function and interaction function. But $\mathcal{U}_{ki}$ are not updated at this stage. They are called supporting learnable parameters. Let $\mathcal{U}_{k}=\{\mathcal{U}_{ki}|i=1,\cdots,M_k\}$ denote the supporting learnable parameters for all active learners. Let $\mathcal{D}_{k}$ denote the active training datasets, consisting of the training dataset of each active learner in $\mathcal{A}_k$.  Let $\mathcal{F}_k$ denote the active auxiliary datasets used in this stage to define the interaction function. The learning activity at stage $k$ is formulated as an optimization problem where the optimization variables are active learnable parameters and the objective involves 1) a training loss $L_k$ defined on the active training datasets, active learnable parameters, and supporting  learnable parameters; 2) (optionally)  an interaction function $I_k$ that depicts the interaction between learners in $\mathcal{A}_k$.  The notations are summarized in Table~\ref{tb:skl_notations}.

\subsubsection{The Mathematical Framework for Skillearn}
The formulation of Skillearn is shown in Eq.(\ref{eq:overall}). 
\begin{equation}
    \begin{array}{ll}
\max_{\{\mathcal{U}_{i}\}_{i=1}^{K}} & L_{val}(\{\mathcal{W}_{j}^*(\{\mathcal{U}_{i}\}_{i=1}^{j})\}_{j=1}^{K},\mathcal{D}^{(\textrm{val})},\mathcal{F})+\gamma_{val} I_{val}(\{\mathcal{W}_{j}^*(\{\mathcal{U}_{i}\}_{i=1}^{j})\}_{j=1}^{K},\mathcal{F}) \quad (\textrm{Validation stage})\\
s.t. 
& \textrm{Learning stage $K$:}\\
&
\mathcal{W}_{K}^*(\{\mathcal{U}_{j}\}_{j=1}^K)=  \underset{\mathcal{W}_{K}}{\textrm{min}}\;  L_K(\mathcal{W}_{K},\mathcal{U}_{K},\{\mathcal{W}_{j}^*(\{\mathcal{U}_{i}\}_{i=1}^{j})\}_{j=1}^{K-1},\mathcal{D}^{(\textrm{tr})}_{K},\mathcal{F}_{K})+\\
&\quad\quad\quad\quad\quad\quad\quad\quad\quad\;\gamma_K I_K(\mathcal{W}_{K},\mathcal{U}_{K},\{\mathcal{W}_{j}^*(\{\mathcal{U}_{i}\}_{i=1}^{j})\}_{j=1}^{K-1},\mathcal{F}_K) \\
&\vdots\\

& \textrm{Learning stage $k$:}\\
 &\mathcal{W}_{k}^*(\{\mathcal{U}_{j}\}_{j=1}^k)= \underset{\mathcal{W}_{k}}{\textrm{min}}\; L_k(\mathcal{W}_{k},\mathcal{U}_{k},\{\mathcal{W}_{j}^*(\{\mathcal{U}_{i}\}_{i=1}^{j})\}_{j=1}^{k-1},\mathcal{D}^{(\textrm{tr})}_{k},\mathcal{F}_{k})+ \\
 &\quad\quad\quad\quad\quad\quad\quad\quad\quad\gamma_k I_k(\mathcal{W}_{k},\mathcal{U}_{k},\{\mathcal{W}_{j}^*(\{\mathcal{U}_{i}\}_{i=1}^{j})\}_{j=1}^{k-1},\mathcal{F}_k) \\

 &\vdots\\
 & \textrm{Learning stage $1$:}\\
  &\mathcal{W}_{1}^*(\mathcal{U}_{1})=  \underset{\mathcal{W}_{1}}{\textrm{min}}\;  L_1(\mathcal{W}_{1},\mathcal{U}_{1},\mathcal{D}^{(\textrm{tr})}_{1}, \mathcal{F}_{1})+ \gamma_1 I_1(\mathcal{W}_{1},\mathcal{U}_{1},\mathcal{F}_1)\\
\end{array}
\label{eq:overall}
\end{equation}
It is a multi-level optimization framework, which involves $K+1$ optimization problems. On the constraints are $K$ optimization problems, each corresponding to a learning stage. The $K$ learning stages are ordered. From bottom to top, the optimization problems correspond to the learning stage $1,2,\cdots,K$ respectively.  In the optimization problem of the learning stage $k$, the optimization variables are the active learnable parameters $\mathcal{W}_k$ of all active learners in this stage. The objective function consists of  a training loss $L_k(\mathcal{W}_{k},\mathcal{U}_{k},\{\mathcal{W}_{j}^*(\{\mathcal{U}_{i}\}_{i=1}^{j})\}_{j=1}^{k-1},\mathcal{D}^{(\textrm{tr})}_{k},\mathcal{F}_{k})$ defined on the active  learnable parameters $\mathcal{W}_k$, supporting learnable parameters $\mathcal{U}_k$, optimal solutions $\{\mathcal{W}_{j}^*(\{\mathcal{U}_{i}\}_{i=1}^{j})\}_{j=1}^{k-1}$ obtained in previous learning stages, active training datasets $\mathcal{D}_k^{(\textrm{tr})}$, and active auxiliary datasets $\mathcal{F}_k$. 
Typically, $L_k(\mathcal{W}_{k},\mathcal{U}_{k},\{\mathcal{W}_{j}^*(\{\mathcal{U}_{i}\}_{i=1}^{j})\}_{j=1}^{k-1},\mathcal{D}^{(\textrm{tr})}_{k},\mathcal{F}_{k})$ can be decomposed into a summation of  active learners' individual training losses:
\begin{equation}
    L_k(\mathcal{W}_{k},\mathcal{U}_{k},\{\mathcal{W}_{j}^*(\{\mathcal{U}_{i}\}_{i=1}^{j})\}_{j=1}^{k-1},\mathcal{D}^{(\textrm{tr})}_{k},\mathcal{F}_{k})=\sum_{i=1}^{M_k} L_{ki}(\mathcal{W}_{ki},\mathcal{U}_{ki},\{\mathcal{W}_{j}^*(\{\mathcal{U}_{i}\}_{i=1}^{j})\}_{j=1}^{k-1},D^{(\textrm{tr})}_{ki},\mathcal{F}_{k})
\end{equation}
where $L_{ki}(\mathcal{W}_{ki},\mathcal{U}_{ki},\{\mathcal{W}_{j}^*(\{\mathcal{U}_{i}\}_{i=1}^{j})\}_{j=1}^{k-1},D^{(\textrm{tr})}_{ki},\mathcal{F}_{k})$ is the training loss of the active learner $i$ defined on its active parameters $\mathcal{W}_{ki}$, supporting parameters $\mathcal{U}_{ki}$, and training dataset $D^{(\textrm{tr})}_{ki}$. The $M_k$ learners do not interact in the training loss. The other part of the objective function is the interaction function $I_k(\mathcal{W}_{k},\mathcal{U}_{k},\{\mathcal{W}_{j}^*(\{\mathcal{U}_{i}\}_{i=1}^{j})\}_{j=1}^{k-1},\mathcal{F}_k)$ which depicts how the $M_k$ active learners interact with each other in this learning stage. It is defined on the active  learnable parameters $\mathcal{W}_k$, supporting learnable parameters $\mathcal{U}_k$, optimal solutions $\{\mathcal{W}_{j}^*(\{\mathcal{U}_{i}\}_{i=1}^{j})\}_{j=1}^{k-1}$ in previous stages, and active auxiliary datasets $\mathcal{F}_k$. $\gamma_k$ is a tradeoff parameter between the training loss and interaction function. $\mathcal{U}_{k}$ is needed to define the objective, but it is not updated at this stage. 
After completing the learning at stage $k$, we obtain the optimal solution  $\mathcal{W}_{k}^*(\{\mathcal{U}_{j}\}_{j=1}^k)$. Note that $\mathcal{W}^*_{k}$ is function of $\{\mathcal{U}_{j}\}_{j=1}^k$ since $\mathcal{W}^*_{k}$ is a function of the objective and the objective is a function of $\{\mathcal{U}_{j}\}_{j=1}^k$. $\mathcal{W}_{k}^*(\{\mathcal{U}_{k}\}_{j=1}^k)$ is used to define the objectives in later stages.

At the very top of Eq.(\ref{eq:overall}), the optimization problem (outside the constraint block) corresponds to the validation stage which validates the optimal solutions $\{\mathcal{W}_{j}^*(\{\mathcal{U}_{i}\}_{i=1}^{j})\}_{j=1}^{K}$ obtained in the $K$ learning stages. The optimization variables are remaining learnable parameters  $\{\mathcal{U}_{i}\}_{i=1}^{K}$ that have not been learned in the $K$ learning stages. The objective function consists of a validation loss and an interaction function. $\gamma_{val}$ is a tradeoff parameter. The validation loss $L_{val}(\{\mathcal{W}_{j}^*(\{\mathcal{U}_{i}\}_{i=1}^{j})\}_{j=1}^{K},\mathcal{D}^{(\textrm{val})},\mathcal{F})$ is defined on the validation sets of all learners $\mathcal{D}^{(\textrm{val})}=\{\mathcal{D}^{(\textrm{val})}_i\}_{i=1}^M$, the optimal solutions $\{\mathcal{W}_{j}^*(\{\mathcal{U}_{i}\}_{i=1}^{j})\}_{j=1}^{K}$, and the auxiliary datasets $\mathcal{F}$. The interaction function is defined on $\{\mathcal{W}_{j}^*(\{\mathcal{U}_{i}\}_{i=1}^{j})\}_{j=1}^{K}$ and  $\mathcal{F}$.

\textbf{Remarks:}
\begin{itemize}
    \item Note that for simplicity, we assume the optimization problem at each stage is a minimization problem. The optimization problem can be more complicated problems such as min-max problems.
    \item At a certain stage, a learnable parameter cannot be simultaneously an active parameter and a supporting parameter. For active parameters in stage $k$, once learned, they cannot be active parameters or supporting parameters in later stages. For supporting parameters in stage $k$, they can be active parameters or supporting parameters in later stages.
    \item The supporting  parameters are not learned in previous stages.
\end{itemize}

\subsection{Optimization Algorithm for Skillearn}
\label{opt:sk}
In this section, we develop an algorithm to solve the Skillearn problem in Eq.(\ref{eq:overall}), inspired by the algorithm in \citep{liu2018darts}. For each learning stage $k$ with an optimization problem: $\mathcal{W}_{k}^*(\{\mathcal{U}_{j}\}_{j=1}^k)=\textrm{min}_{\mathcal{W}_{k}}  L_k(\mathcal{W}_{k},\mathcal{U}_{k},\{\mathcal{W}_{j}^*(\{\mathcal{U}_{i}\}_{i=1}^{j})\}_{j=1}^{k-1},\mathcal{D}^{(\textrm{tr})}_{k},\mathcal{F}_{k})+\gamma_k I_k(\mathcal{W}_{k},\mathcal{U}_{k},\{\mathcal{W}_{j}^*(\{\mathcal{U}_{i}\}_{i=1}^{j})\}_{j=1}^{k-1},\mathcal{F}_k)$, we approximate the optimal solution  $\mathcal{W}_{k}^*(\{\mathcal{U}_{j}\}_{j=1}^k)$ by one-step gradient descent update of the variable $\mathcal{W}_{k}$:
\begin{equation}
\begin{array}{l}
    \mathcal{W}_{k}^*(\{\mathcal{U}_{j}\}_{j=1}^k)\approx \mathcal{W}'_{k}(\{\mathcal{U}_{j}\}_{j=1}^k)=\\
    \mathcal{W}_{k}-\eta \nabla_{\mathcal{W}_{k}}(L_k(\mathcal{W}_{k},\mathcal{U}_{k},\{\mathcal{W}_{j}^*(\{\mathcal{U}_{i}\}_{i=1}^{j})\}_{j=1}^{k-1},\mathcal{D}^{(\textrm{tr})}_{k},\mathcal{F}_{k})+\gamma_k I_k(\mathcal{W}_{k},\mathcal{U}_{k},\{\mathcal{W}_{j}^*(\{\mathcal{U}_{i}\}_{i=1}^{j})\}_{j=1}^{k-1},\mathcal{F}_k)).
    \end{array}
\end{equation}
In learning stages $k+1,\cdots,K$,  $\mathcal{W}_{k}^*(\{\mathcal{U}_{j}\}_{j=1}^k)$ may be used to define objective functions. For a stage $l$ where $k<l<K$, if its objective involves $\mathcal{W}_{k}^*(\{\mathcal{U}_{j}\}_{j=1}^k)$, we replace $\mathcal{W}_{k}^*(\{\mathcal{U}_{j}\}_{j=1}^k)$ with $\mathcal{W}'_{k}(\{\mathcal{U}_{j}\}_{j=1}^k)$ and get an approximated objective. When approximating  $\mathcal{W}_{l}^*(\{\mathcal{U}_{j}\}_{j=1}^l)$, we use the gradient of the approximated objective:
\begin{equation}
\begin{array}{l}
    \mathcal{W}_{l}^*(\{\mathcal{U}_{j}\}_{j=1}^l)\approx \mathcal{W}'_{l}(\{\mathcal{U}_{j}\}_{j=1}^l)=\\
    \mathcal{W}_{l}-\eta \nabla_{\mathcal{W}_{l}}(L_l(\mathcal{W}_{l},\mathcal{U}_{l},\{\mathcal{W}'_{j}(\{\mathcal{U}_{i}\}_{i=1}^{j})\}_{j=1}^{l-1},\mathcal{D}^{(\textrm{tr})}_{l},\mathcal{F}_{l})+\gamma_l I_l(\mathcal{W}_{l},\mathcal{U}_{l},\{\mathcal{W}'_{j}(\{\mathcal{U}_{i}\}_{i=1}^{j})\}_{j=1}^{l-1},\mathcal{F}_l)).
    \end{array}
    \label{eq:w_ln_appro}
\end{equation}
For the objective in the validation stage, it can be approximated as:
\begin{equation}
    L_{val}(\{\mathcal{W}'_{j}(\{\mathcal{U}_{i}\}_{i=1}^{j})\}_{j=1}^{K},\mathcal{D}^{(\textrm{val})},\mathcal{F})+\gamma_{val} I_{val}(\{\mathcal{W}'_{j}(\{\mathcal{U}_{i}\}_{i=1}^{j})\}_{j=1}^{K},\mathcal{F}).
    \label{eq:val_appro}
\end{equation}
We update the remaining learnable parameters $\{\mathcal{U}_{i}\}_{i=1}^{K}$ by minimizing this approximated objective. The optimization algorithm for Skillearn is summarized in Algorithm~\ref{algo:algo-sk}.   \begin{algorithm}[H]
\SetAlgoLined
 \While{not converged}{
1. For $k=1\cdots K$, update $\mathcal{W}_{k}^*(\{\mathcal{U}_{j}\}_{j=1}^k)$ using Eq.(\ref{eq:w_ln_appro})\\
2. Update $\{\mathcal{U}_{i}\}_{i=1}^{K}$  by minimizing the approximated objective in Eq.(\ref{eq:val_appro})
 }
 \caption{Optimization algorithm for Skillearn}
 \label{algo:algo-sk}
\end{algorithm}

\section{Case Study I: Learning by Passing Tests}
In this section, we apply our general Skillearn framework to formalize a human learning technique -- learning by passing tests, and apply it to improve machine learning. 
In human learning, an effective and widely used methodology for improving learning outcome is to let the learner take increasingly more-difficult tests. To successfully pass a more challenging test, the learner needs to gain better learning ability. By progressively passing tests that have increasing levels of difficulty, the learner strengthens his/her learning capability gradually.

Inspired by this test-driven learning technique of humans, we are interested in investigating whether this methodology is helpful for improving machine learning as well. We use the Skillearn framework to formalize this human learning technique, which results in a novel machine learning framework called learning by passing tests (LPT). In this framework, there are two learners: a ``testee" model and a ``tester" model. 
The tester creates a sequence of ``tests" with growing levels of difficulty. The testee tries to learn better  so that it can pass these increasingly more-challenging tests. Given a large collection of data examples called ``test bank", the tester   creates a test $T$ by selecting a subset of examples from the test bank. The testee applies its intermediately-trained model $M$ to make predictions on the examples in $T$. The prediction error rate $R$ reflects how difficult this test is. If the testee can make correct predictions on $T$, it means that $T$ is not difficult enough. The tester will create a more challenging test $T'$ by selecting a new set of examples from the test bank in a way that the new error rate $R'$ achieved by $M$ is larger than $R$. Given this more demanding test $T'$, the testee re-learns its model to pass $T'$, in a way that the newly-learned model $M'$ achieves a new error rate  $R''$ on $T'$ where $R''$ is smaller than $R'$. This process iterates until convergence.

In our framework, both the testee and tester perform learning. The testee learns how to best conduct a target task $J_1$ and the tester learns how to create difficult and meaningful tests. To encourage a created test $T$ to be meaningful, the tester trains a model using $T$ to perform a target task $J_2$. If the model performs well on $J_2$, it indicates that $T$ is meaningful. 
The testee has two sets of learnable parameters: neural architecture and network weights. The tester has three learnable modules: data encoder, test creator, and target-task executor. The learning is organized into three stages. In the first stage, the testee trains its network weights on the training set of task $J_1$ with the architecture fixed. In the second stage, the tester trains its data encoder and target-task executor on a created test to perform the target task $J_2$, with the test creator fixed. In the third stage, the testee updates its model architecture by minimizing the predictive loss $L$ on the test created by the tester; the tester updates its test creator by maximizing $L$ and minimizing the loss on the validation set of $J_2$. The testee and tester interact on the loss function $L$ in an adversarial manner, where the testee minimizes this loss while the tester maximizes this loss. 
The three stages are performed jointly end-to-end in a multi-level optimization framework, where a latter stage  influences an earlier stage and vice versa. We apply our method for neural architecture search~\citep{zoph2016neural,liu2018darts,real2019regularized} in image  classification tasks on CIFAR-100, CIFAR-10, and ImageNet~\citep{deng2009imagenet}. Our method achieves significant improvement over state-of-the-art baselines.

\subsection{Method}
In this section, we describe how to instantiate the general Skillearn framework to the LPT framework, and how to instantiate the general optimization procedure of Skillearn to a specialized optimization algorithm for LPT.

\subsubsection{Learning by Passing Tests}
In the learning by passing tests (LPT) framework, there are two learners: a testee model and a tester model, where the testee studies how to perform a target task $J_1$ such as classification, regression, etc. The eventual goal is to make the testee achieve a better learning outcome with the help of the tester. There is a collection of data examples called ``test bank". The tester creates a test by selecting a subset of examples from the test bank. Given a test $T$, the testee applies its intermediately-trained model $M$ to make predictions on $T$ and measures the prediction error rate $R$. From the perspective of the tester, $R$ indicates how difficult the test $T$ is. If $R$ is small, it means that the testee  can easily pass this test. Under such circumstances, the tester will create a more difficult test $T'$ which renders the new error rate $R'$ achieved by $M$ on  $T'$ is larger than $R$. From the testee's perspective, $R'$ indicates how well the testee performs on the test. Given this more difficult test $T'$, the testee refines its model to pass this new test. It aims to learn a new model $M'$ in a way that the error rate $R''$ achieved by  $M'$ on $T'$ is smaller than $R'$. This process iterates until an equilibrium is reached. In addition to being difficult, the created test should be meaningful as well. It is possible that the test bank contains poor-quality examples where the class labels may be incorrect or the input data instances are outliers. Using an unmeaningful test containing poor-quality examples to guide the learning of the testee may render the testee to overfit these bad-quality examples and generalize poorly on unseen data. To address this problem, we encourage the tester to generate meaningful tests by leveraging the generated tests to perform a target task $J_2$ (e.g., classification). Specifically, the tester uses  examples in the test to train a model for performing $J_2$. If the performance (e.g., accuracy) $P$ achieved by this model in conducting $J_2$ is high, the test is considered to be meaningful. The tester aims to create a test that can yield a high $P$.

\begin{figure}[t]
    \centering
 \includegraphics[width=0.8\textwidth]{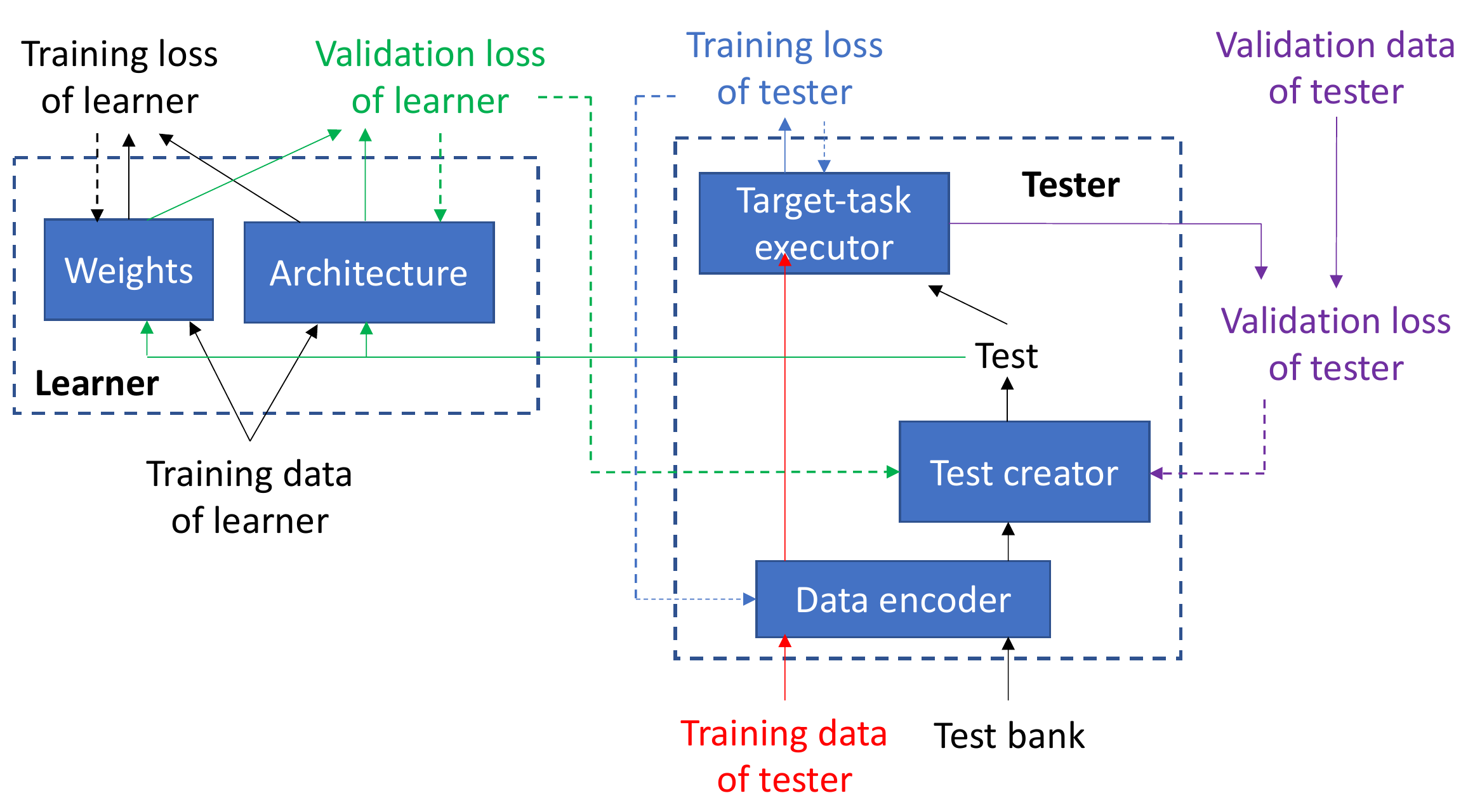}
       \caption{Illustration of learning by passing tests. The solid arrows denote the process of making predictions and calculating losses. The dotted arrows denote the process of updating learnable parameters by minimizing corresponding losses.}
 \label{fig:arch}
\end{figure}

\begin{table}[t]
\centering
\begin{tabular}{l|l}
\hline
Notation & Meaning \\
\hline
$A$ & Architecture of the testee\\
$W$ & Network weights of the testee\\
$E$ & Data encoder of the tester\\
$C$ & Test creator of the tester \\
$X$ & Target-task executor of the tester\\
$D_{ee}^{(\textrm{tr})}$ & Training data of the testee\\
$D_{er}^{(\textrm{tr})}$ & Training data of the tester\\
$D_{er}^{(\textrm{val})}$ & Validation data of the tester\\
$D_b$ & Test bank \\
\hline
\end{tabular}
\caption{Notations in Learning by Passing Tests}
\label{tb:notations}
\end{table}

In our framework, both the testee and the tester performs learning. The testee studies how to best fulfill the target task $J_1$. The tester studies how to create tests that are difficult and meaningful.  In the testee' model, there are two sets of learnable parameters: model architecture  and network weights. The architecture and weights are both used to make predictions in $J_1$.
The tester's model performs two tasks simultaneously: creating tests and performing target-task $J_1$. The model has three  modules with learnable parameters: data encoder, test creator, and target-task executor, where the test creator performs the task of generating tests and the target-task executor conducts $J_1$.  The test creator and target-task executor share the same data encoder. The data encoder takes a data example $d$ as input and generates a latent representation for this example. Then the representation is fed into the test creator which determines whether $d$ should be selected into the test. The  representation is also fed into the target-task executor which performs prediction on $d$ during performing the target task $J_2$.

\begin{table}[t]
\centering
\begin{tabular}{l|p{8cm}}
\hline
Active learners & Testee\\
\hline
Active learnable parameters & Network weights of the testee\\
\hline
Supporting learnable parameters & Architecture of the testee\\
\hline
Active training datasets & Training dataset of target-task $J_1$ performed by the testee \\
\hline
Active auxiliary datasets & -- \\
\hline
Training loss & Training loss of target-task $J_1$: $L(A, W, D_{ee}^{(\mathrm{tr})})$   \\
\hline
Interaction function & -- \\
\hline
Optimization problem & $W^{*}(A)=\min _{W} L(A, W, D_{ee}^{(\mathrm{tr})})$ \\
\hline
\end{tabular}
\caption{Learning Stage I in LPT}
\label{tb:lpt-s1}
\end{table}

\begin{table}[t]
\centering
\begin{tabular}{l|p{8cm}}
\hline
Active learners & Examiner\\
\hline
Active learnable parameters & 1) Data encoder of the tester; 2) Target-task executor of the tester.  \\
\hline
Supporting learnable parameters & Test creator of the tester\\
\hline
Active training datasets & Training data of target-task $J_2$ performed by the tester\\
\hline
Active auxiliary datasets & Test bank\\
\hline
Training loss & $L(E, X, D_{er}^{(\mathrm{tr})}) +\gamma L(E, X, \sigma(C, E, D_{b}))$\\
\hline
Interaction function & -- \\
\hline
Optimization problem & $    E^{*}(C), X^{*}(C)=\min _{E, X} \;\; L(E, X, D_{er}^{(\mathrm{tr})}) +\gamma L(E, X, \sigma(C, E, D_{b})).$\\
\hline
\end{tabular}
\caption{Learning Stage II in LPT }
\label{tb:lpt-s2}
\end{table}

In our framework, the learning of the testee and the tester is organized into three stages. In the first stage, the testee learns its network weights $W$ by minimizing the training loss $L(A, W, D_{ee}^{(\mathrm{tr})})$ defined on the training data $D_{ee}^{(\mathrm{tr})}$ in the  task $J_1$. The architecture $A$ is used to define the training loss, but it is not learned in this stage. If $A$ is learned by minimizing this training loss, a trivial solution will be yielded where $A$ is very large and complex that it can perfectly overfit the training data but will generalize poorly on unseen data. Let $W^*(A)$ denotes the optimally learned $W$ in this stage. 
Note that $W^*$ is a function of $A$ because $W^*$ is a function of the training loss and the training loss is a function of $A$. Table~\ref{tb:lpt-s1} shows the key elements of this learning stage under the Skillearn terminology. The testee is the active learner, which performs learning in this stage. Network weights of the testee are the active learnable parameters, which are updated at this stage. The architecture variables of the testee are the supporting learnable parameters, which are used to define the loss function, but are not updated at this stage. Active training datasets include the training data of the task $J_1$ performed by the testee. There are no active auxiliary datasets. Training loss is $L(A, W, D_{ee}^{(\mathrm{tr})})$. There is no interaction function at this stage. The optimization problem is:
\begin{equation}
    W^{*}(A)=\min _{W} L\left(A, W, D_{ee}^{(\mathrm{tr})}\right).
\end{equation}

In the second stage, the tester learns its data encoder $E$ and target-task executor $X$ by minimizing the training loss $L(E, X, D_{er}^{(\mathrm{tr})}) +\gamma L(E, X, \sigma(C, E, D_{b}))$ in the task $J_2$. The training loss consists  of two parts. The first part $L(E, X, D_{er}^{(\mathrm{tr})})$ is defined on the training dataset $D_{er}^{(\textrm{tr})}$ in $J_2$. The second part $L(E, X, \sigma(C, E, D_{b}))$ is defined on the test $\sigma(C, E, D_{b})$ created  by the test creator. For each example $d$ in the test bank $D_{b}$, it is first fed into the encoder $E$, then the creator $C$, which outputs a binary value  indicating whether $d$ should be selected into the test. $\sigma(C, E, D_{b})$ is the collection of examples whose binary value is equal to 1. $\gamma$ is a tradeoff parameter between these two parts of losses. The creator $C$ is used to define the second-part loss, but it is not learned in this stage. Otherwise, a trivial solution will be yielded where $C$ always sets the binary value to 0 for each test-bank example so that the second-part loss becomes 0. Let $E^*(C)$ and $X^*(C)$ denote the optimally trained $E$ and $X$ in this stage. Note that they are both functions of $C$ since they are functions of the training loss and the training loss is a function of $C$. Table~\ref{tb:lpt-s2} shows the key elements of this learning stage under the Skillearn terminology. The tester is the active learner. The active learnable parameters include the data encoder and target-task executor of the tester. The supporting learnable parameters include the test creator. The active training datasets include the training data of target-task $J_2$ performed by the tester. The active auxiliary datasets include the test bank. The training loss is $L(E, X, D_{er}^{(\mathrm{tr})}) +\gamma L(E, X, \sigma(C, E, D_{b}))$. There is no interaction function at this stage. The optimization problem is:
\begin{equation}
    E^{*}(C), X^{*}(C)=\min _{E, X} \;\; L\left(E, X, D_{er}^{(\mathrm{tr})}\right) +\gamma L\left(E, X, \sigma\left(C, E, D_{b}\right)\right).
\end{equation}

\begin{table}[t]
\centering
\begin{tabular}{p{4cm}|p{10.5cm}}
\hline
Active learners & Testee, tester\\
\hline
Remaining learnable parameters & 1) Architecture of the testee; 2) Test creator of the tester\\
\hline
Validation datasets& Validation dataset of the tester\\
\hline
Active auxiliary datasets & Test bank\\
\hline
Validation loss & $L(E^{*}(C), X^{*}(C), D_{er}^{(\mathrm{val})})$\\
\hline
Interaction function &  Testee's prediction loss defined on the test created by the tester: $ L(A, W^{*}(A), \sigma(C, E^{*}(C), D_{b}))/|\sigma(C, E^{*}(C), D_{b})|$\\
\hline
Optimization problem &  $\max_{C} \min _{A}\;\;  L(A, W^{*}(A), \sigma(C, E^{*}(C), D_{b}))/|\sigma(C, E^{*}(C), D_{b})|-\lambda L(E^{*}(C), X^{*}(C), D_{er}^{(\mathrm{val})})$\\
\hline
\end{tabular}
\caption{Validation Stage in LPT}
\label{tb:lpt-vs}
\end{table}

In the third stage, the testee learns its architecture by trying to pass the test $\sigma(C, E^*(C), D_{b})$ created by the tester. Specifically, the testee aims to minimizes the predictive loss  of its model on the test:
\begin{equation}
    L(A, W^{*}(A), \sigma(C, E^{*}(C), D_{b}))=\sum_{d\in \sigma(C, E^{*}(C), D_{b}) } \ell(A, W^{*}(A), d)
\end{equation}
where $d$ is an example in the test and $\ell(A, W^{*}(A), d)$ is the loss defined in this example. 
A smaller $L(A, W^{*}(A), \sigma(C, E^{*}(C), D_{b}))$ indicates that the testee performs well on this test. Meanwhile, the tester learns its test creator $C$ in a way that $C$ can create a test with more difficulty and meaningfulness. Difficulty is measured by the testee's predictive loss $L(A, W^{*}(A), \sigma(C, E^{*}(C), D_{b}))$ on the test. Given a model $(A, W^{*}(A))$ of the testee and two tests of the same size (same number of examples): $\sigma(C_1, E^{*}(C_1), D_{b})$ created by $C_1$ and $\sigma(C_2, E^{*}(C_2), D_{b})$ created by $C_2$, if $L(A, W^{*}(A), \sigma(C_1, E^{*}(C_1), D_{b}))> L(A, W^{*}(A), \sigma(C_2, E^{*}(C_2), D_{b}))$, it means that $\sigma(C_1, E^{*}(C_1), D_{b})$ is more challenging to pass than $\sigma(C_2, E^{*}(C_2), D_{b})$.   
Therefore, the tester can learn to create a more challenging test by maximizing $L(A, W^{*}(A), \sigma(C, E^{*}(C), D_{b}))$. A trivial solution of increasing $L(A, W^{*}(A), \sigma(C, E^{*}(C), D_{b}))$ is to enlarge the size of the test. But a larger size does not imply more difficulty. To discourage this degenerated solution from happening, we normalize the loss using the size of the test: 
\begin{equation}
    \frac{1}{\left|\sigma\left(C, E^{*}(C), D_{b}\right)\right|} L\left(A, W^{*}\left(A\right), \sigma\left(C, E^{*}(C), D_{b}\right)\right)
    \label{eq:interact}
\end{equation}
where $|\sigma(C, E^{*}(C), D_{b})|$ is the cardinality of the set $\sigma(C, E^{*}(C), D_{b})$. Under the Skillearn terminologies, the loss in Eq.(\ref{eq:interact}) is the interaction function where the testee and tester interact. The testee aims to minimize this loss to ``pass" the testee and the tester aims to maximize this loss to ``fail" the testee. 
To measure the meaningfulness of a test, we check how well the optimally-trained task executor $E^*(C)$ and data encoder $X^*(C)$ of the tester  perform on the validation data $D_{er}^{\textrm{(val)}}$ in the target task $J_2$, and the performance is measured by the validation loss: $L(E^{*}(C), X^{*}(C), D_{er}^{(\mathrm{val})})$. $E^*(C)$ and $X^*(C)$ are trained using the test generated by $C$ in the second stage. If the validation loss is small, it means that the created test is helpful in training the task executor and therefore is considered as being meaningful. To create a meaningful test, the tester learns $C$ by minimizing  $L(E^{*}(C), X^{*}(C), D_{er}^{(\mathrm{val})})$. In sum, $C$ is learned by maximizing $L(A, W^{*}(A), \sigma(C, E^{*}(C), D_{b}))/|\sigma(C, E^{*}(C), D_{b})|-\lambda L(E^{*}(C), X^{*}(C), D_{er}^{(\mathrm{val})})$, where $\lambda$ is a tradeoff parameter between these two objectives. Under the Skillearn terminology, this stage is a validation stage. Table~\ref{tb:lpt-vs} summarizes the key elements of this stage. 
The active learners include both the testee and the tester. The remaining learnable parameters include the architecture of the testee and the test creator of the tester. The validation datasets include the validation data in the target-task $J_2$ performed by the tester. The active auxiliary datasets include the test bank. The validation loss is $L(E^{*}(C), X^{*}(C), D_{er}^{(\mathrm{val})})$. The interaction function is $\frac{1}{|\sigma(C, E^{*}(C), D_{b})|} L(A, W^{*}(A), \sigma(C, E^{*}(C), D_{b}))$. The optimization problem is:
\begin{equation}
    \max _{C} \min _{A}\;\; \frac{1}{\left|\sigma\left(C, E^{*}(C), D_{b}\right)\right|} L\left(A, W^{*}\left(A\right), \sigma\left(C, E^{*}(C), D_{b}\right)\right)-\lambda L\left(E^{*}(C), X^{*}(C), D_{er}^{(\mathrm{val})}\right).
\end{equation}

\begin{table}[t]
\centering
\begin{tabular}{l|p{10.5cm}}
\hline
Skillearn & Learning by Passing Tests\\
\hline 
Learners & 1) Testee; 2) Tester \\
\hline
Learnable parameters &  1) Architecture of testee; 2) Network weights of testee; 3) Data encoder of tester; 4) Target-task executor of tester; 5) Test creator of tester. \\
\hline
Interaction function & Testee's prediction loss defined on the test created by the tester: $ L(A, W^{*}(A), \sigma(C, E^{*}(C), D_{b}))/|\sigma(C, E^{*}(C), D_{b})|$ \\
\hline
Learning stages &  Learning stage I: the testee learns its network weights on its training data: $W^{*}(A)=\min _{W} L(A, W, D_{ee}^{(\mathrm{tr})})$\newline
Learning stage II: the tester uses its test creator to select a subset of examples from the test bank, then it learns its data encoder and target-task executor on its training data and on the selected examples from the test bank: $E^{*}(C), X^{*}(C)=\min _{E, X} \;\; L(E, X, D_{er}^{(\mathrm{tr})}) +\gamma L(E, X, \sigma(C, E, D_{b})).$
\\
\hline 
Validation stage & 1) The testee updates its architecture to minimize the prediction loss on the test created by the tester; 2) The tester updates its test creator to maximize the testee's prediction loss and minimize its own validation loss. $\max _{C} \min _{A}\;\; \frac{1}{|\sigma(C, E^{*}(C), D_{b})|} L(A, W^{*}(A), \sigma(C, E^{*}(C), D_{b}))-\lambda L(E^{*}(C), X^{*}(C), D_{er}^{(\mathrm{val})}).$\\
\hline 
Datasets & 1) Training data of the testee; 2) Training data of the tester; 3) Validation data of the tester; 4) Test bank.\\
\hline
\end{tabular}
\caption{Instantiation of Skillearn to LPT}
\label{tb:sltolpt}
\end{table}

The three stages are mutually dependent: $W^*(A)$ learned in the first stage and $E^*(C)$ and $X^*(C)$ learned in the second stage are used to define the objective function in the third stage; the updated $C$ and $A$ in the third stage in turn change the objective functions in the first and second stage, which subsequently render $W^*(A)$, $E^*(C)$, and $X^*(C)$ to be changed. 
Putting these pieces together, we instantiate the Skillearn framework into the following LPT formulation:
\begin{equation}
    \begin{array}{l}
\max _{C} \min _{A}\;\; \frac{1}{\left|\sigma\left(C, E^{*}(C), D_{b}\right)\right|} L\left(A, W^{*}\left(A\right), \sigma\left(C, E^{*}(C), D_{b}\right)\right)-\lambda L\left(E^{*}(C), X^{*}(C), D_{er}^{(\mathrm{val})}\right) \textrm{(III)} \\
s.t. \;\;   E^{*}(C), X^{*}(C)=\min _{E, X} \;\; L\left(E, X, D_{er}^{(\mathrm{tr})}\right) +\gamma L\left(E, X, \sigma\left(C, E, D_{b}\right)\right) \textrm{(Stage II)} \\
\quad\;\;\;
W^{*}\left(A\right)=\min _{W}\;\; L\left(A, W, D_{ee}^{(\mathrm{tr})}\right) \textrm{(Stage I)}
\end{array}
\label{eq:learning_objective}
\end{equation}
This formulation nests three optimization problems. On the constraints of the outer optimization problem are two inner optimization problems corresponding to the first and second learning stage respectively. The objective function of the outer optimization problem corresponds to the validation stage. Table~\ref{tb:sltolpt} summarizes  the instantiation of Skillearn to LPT.

As of now, the test $\sigma(C, E, D_{b})$ is represented as a subset, which is highly discrete and therefore difficult for optimization. To address this problem, we perform a continuous relaxation of $\sigma(C, E, D_{b})$:
\begin{equation}
   \sigma(C, E, D_{b}) =\{(d,f(d,C,E))|d\in D_{b}\}
\end{equation}
where for each example $d$ in the test bank, the original binary value indicating whether $d$ should be selected is now relaxed to a continuous probability $f(d,C,E)$ representing how likely $d$ should be selected. Under this relaxation, $L(E, X, \sigma(C, E, D_{b}))$ can be computed as follows:
\begin{equation}
    L(E, X, \sigma(C, E, D_{b}))= \sum_{d\in D_{b}} f(d,C,E) \ell (E,X,d)
\end{equation}
where we calculate the loss $\ell (E,X,d)$ on each test-bank example and weigh this loss using $f(d,C,E))$. If $f(d,C,E))$ is small, it means that $d$ is less likely to be selected into the test and its corresponding loss should be down-weighted. Similarly, $L(A, W^{*}(A), \sigma(C, E^{*}(C), D_{b}))$ is calculated as $\sum_{d\in D_{b}} f(d,C,E^{*}(C)) \ell (A, W^{*}(A),d)$. And $|\sigma(C, E^{*}(C), D_{b})|$ can be calculated as
\begin{equation}
    |\sigma(C, E^{*}(C), D_{b})|=\sum_{d\in D_{b}} f(d,C,E^{*}(C))
\end{equation}
Similar to \citep{liu2018darts}, we represent the architecture $A$ of the testee in a differentiable way. The search space of $A$ is composed of a large number of building blocks. The output of each block is associated with a variable $a$ indicating how important this block is. After learning, blocks whose $a$ is among the largest are retained to form the final architecture. In this end, architecture search amounts to optimizing the set of architecture variables $A=\{a\}$.

\subsubsection{Optimization Algorithm}
In this section, we instantiate the general optimization framework in Section~\ref{opt:sk} to derive an optimization algorithm for LPT.  
We approximate $E^{*}(C)$ and $X^{*}(C)$ using one-step gradient descent update of $E$ and $X$ with respect to $L(E, X, D_{er}^{(\mathrm{tr})}) +\gamma L(E, X, \sigma(C, E, D_{b}))$ and approximate $W^{*}(A)$ using one-step gradient descent update of $W$  with respect to $L(A, W, D_{ee}^{(\mathrm{tr})})$. Then we plug in these approximations into \begin{equation}
 L(A, W^{*}(A), \sigma(C, E^{*}(C), D_{b}))/|\sigma(C, E^{*}(C), D_{b})|-\lambda L(E^{*}(C), X^{*}(C), D_{er}^{(\mathrm{val})}),
 \label{eq:3rd-obj}
\end{equation}
and perform gradient-descent update of $C$ and $A$ with respect to this approximated objective. In the sequel, we use $\nabla^2_{Y,X}f(X,Y)$ to denote $\frac{\partial f(X,Y)}{\partial X\partial Y}$.

Approximating $W^{*}(A)$ using $W'=W - \xi_{ee}  \nabla_{W}L(A, W, D_{ee}^{(\mathrm{tr})})$ where $\xi_{ee}$ is a learning rate and simplifying the notation of $ \sigma(C, E^*(C), D_{b})$ as $\sigma$, we can calculate the approximated gradient of $L\left(A, W^{*}\left(A\right),\sigma\right)$  w.r.t $A$ as:
\begin{equation}
\begin{array}{l}
     \nabla_{A} L\left(A, W^{*}\left(A\right),\sigma\right)\approx  \\
     \nabla_{A}  L\left(A,W - \xi_{ee}  \nabla_{W}L\left(A, W, D_{ee}^{(\mathrm{tr})}\right), \sigma\right)=\\
    \nabla_{A} L\left(A, W^{\prime}, \sigma\right)-\xi_{ee} \nabla_{A, W}^{2} L\left(A, W, D_{ee}^{(\mathrm{tr})}\right) \nabla_{W^{\prime}} L\left(A, W^{\prime},\sigma\right),
\end{array}
\label{eq:descent_arch}
\end{equation}
The second term in the third line   involves expensive matrix-vector product, whose computational complexity can be reduced by a finite difference approximation:
\begin{equation}
\begin{array}{ll}
     \nabla_{A, W}^{2} L\left(A, W, D_{ee}^{(\mathrm{tr})}\right)\nabla_{W^{\prime}} L\left(A, W^{\prime},\sigma\right)\approx  
     \frac{1}{2\alpha_{ee}}\left(\nabla_{A} L\left(A, W^{+}, D_{ee}^{(\mathrm{tr})}\right)-\nabla_{A} L\left(A, W^{-}, D_{ee}^{(\mathrm{tr})}\right)\right),
\end{array}
\label{eq:finite-aw}
\end{equation}
where $W^{\pm}=W \pm \alpha_{ee} \nabla_{W^{\prime}} L\left(A, W^{\prime},\sigma\right)$ and $\alpha_{ee}$ is a small scalar that equals $0.01 /\left\|\nabla_{W^{\prime}} L\left(A, W^{\prime},\sigma\right))\right\|_{2}$.
We approximate $E^*(C)$ and $X^*(C)$ using the following one-step gradient descent update of $E$ and $C$ respectively:
\begin{equation}
\begin{array}{l}
    E^{\prime}=E-\xi_{E} \nabla_{E}[ L(E, X, D_{er}^{(\mathrm{tr})})+\gamma L(E, X, \sigma(C,E,D_b))]\\
    
    X^{\prime}=X-\xi_{X} \nabla_{X}[ L(E, X, D_{er}^{(\mathrm{tr})})+\gamma L(E, X, \sigma(C,E,D_b))]
    \label{eq:update_ec}
    \end{array}
\end{equation}
where $\xi_{E}$ and $\xi_{X}$ are learning rates. Plugging in these approximations into the objective function in Eq.(\ref{eq:3rd-obj}), we can learn $C$ by maximizing the following objective using gradient methods:
\begin{equation}
    L(A, W^{\prime}, \sigma(C, E^{\prime}, D_{b}))/|\sigma(C, E', D_{b})|-\lambda L(E^{\prime}, X^{\prime}, D_{er}^{(\mathrm{val})})
\end{equation}
The derivative of the second term in this objective with respect to $C$ can be calculated as:
\begin{equation}
\begin{array}{l}
\nabla _{C}L(E^{\prime}, X^{\prime}, D_{er}^{(\mathrm{val})})=
\frac{\partial E'}{\partial C} \nabla _{E'} L(E^{\prime}, X^{\prime}, D_{er}^{(\mathrm{val})}) +
\frac{\partial X'}{\partial C}\nabla _{X'} L(E^{\prime}, X^{\prime}, D_{er}^{(\mathrm{val})})\\
\end{array}
\label{eq:grad_c}
\end{equation}
where 
\begin{equation}
\begin{array}{l}
    \frac{\partial E'}{\partial C}=-\xi_{E}\gamma \nabla^{2}_{C,E} L(E, X, \sigma(C,E,D_b))\\
    \frac{\partial X'}{\partial C}=-\xi_{X}\gamma \nabla^{2}_{C,X} L(E, X, \sigma(C,E,D_b))\\
    \end{array}
    \label{eq:sec-gra-ec}
\end{equation}
Similar to Eq.(\ref{eq:finite-aw}), using finite difference approximation to calculate $\nabla^{2}_{C,E} L(E, X, \sigma(C,E,D_b))$\\$\nabla _{E'} L(E^{\prime}, X^{\prime}, D_{er}^{(\mathrm{val})})$ and $\nabla^{2}_{C,X} L(E, X, \sigma(C,E,D_b))\nabla _{X'} L(E^{\prime}, X^{\prime}, D_{er}^{(\mathrm{val})})$, we have:
\begin{equation}
\begin{array}{l}
    \nabla _{C}L(E^{\prime}, X^{\prime}, D_{er}^{(\mathrm{val})})=\\
    -\gamma\xi_{E}\frac{\nabla_{C}L(E^+,X,\sigma(C,E^+,D_b))-\nabla_{C}L(E^-,X,\sigma(C,E^-,D_b))}{2\alpha_{E}}
    -\gamma\xi_{X}\frac{\nabla_{C}L(E,X^+,\sigma(C,E,D_b))-\nabla_{C}L(E,X^-,\sigma(C,E,D_b))}{2\alpha_{X}}
    \end{array}
\end{equation}
where $E^{\pm}=E\pm\alpha_{E} \nabla_{E^\prime}L(E^\prime,X^\prime,D_{er}^{\mathrm{(val)}})$ and $X^{\pm}=X\pm\alpha_{X} \nabla_{X^\prime}L(E^\prime,X^\prime,D_{er}^{\mathrm{(val)}})$. 
For the first term $L(A, W^{\prime}, \sigma(C, E^{\prime}, D_{b}))/|\sigma(C, E', D_{b})|$ in the objective, we can use chain rule to calculate its derivative w.r.t $C$, which involves calculating the derivative of $L(A, W^{\prime}, \sigma(C, E^{\prime}, D_{b}))$ and $|\sigma(C, E', D_{b})|$ w.r.t to $C$. 
The derivative of $L(A, W^{\prime}, \sigma(C, E^{\prime}, D_{b}))$ w.r.t $C$ can be calculated as:
\begin{equation}
\begin{array}{l}
 \nabla _{C}L(A, W^{\prime}, \sigma(C, E^{\prime}, D_{b}))=
 \frac{\partial E'}{\partial C } \nabla_{E^{\prime}}L(A, W^{\prime}, \sigma(C, E^{\prime}, D_{b})),
\end{array}
\label{eq:descent_teach_v2}
\end{equation}
where $ \frac{\partial E'}{\partial C }$ is given in Eq.(\ref{eq:sec-gra-ec}) and $ \nabla^{2}_{C,E} L(E, X, \sigma(C,E,D_b))$
$\times \nabla_{E^{\prime}}L(A, W^{\prime}, \sigma(C, E^{\prime}, D_{b}))$ can be approximated with $\frac{1}{2\alpha_{E}}(\nabla_{C}L(E^+,X,\sigma(C,E^+,D_b))-\nabla_{C}L(E^-,X,\sigma(C,E^-,D_b)))$,\\ where $E^{\pm}$ is $E\pm\alpha_{E}\nabla_{E^{\prime}}L(A, W^{\prime}, \sigma(C, E^{\prime}, D_{b}))$. The derivative of $|\sigma(C, E', D_{b})|=\sum_{d\in D_{b}} f(d,C,E')$ w.r.t $C$ can be calculated as
\begin{equation}
    \sum_{d\in D_{b}} \nabla_C f(d,C,E')+\frac{\partial E'}{\partial C} \nabla_{E'} f(d,C,E')
    \label{eq:grad_cardi}
\end{equation}
where $\frac{\partial E'}{\partial C}$ is given in Eq.(\ref{eq:sec-gra-ec}). The algorithm for solving LPT is summarized in Algorithm~\ref{algo:algo}.

\begin{algorithm}[H]
\SetAlgoLined
 \While{not converged}{
1. Update the architecture of the  testee  by descending  the gradient calculated in Eq.(\ref{eq:descent_arch})\\
2. Update the test creator of the tester by ascending the gradient calculated in Eq.(\ref{eq:grad_c}-\ref{eq:grad_cardi})\\
3. Update the data encoder and target-task executor of the tester using Eq.(\ref{eq:update_ec})\\
4. Update the weights of the testee  by descending $\nabla_{W}L(A, W, D_{ee}^{(\mathrm{tr})})$
 }
 \caption{Optimization algorithm for learning by passing tests}
 \label{algo:algo}
\end{algorithm}

\subsection{Experiments}
\label{sec:exp_lpt}
We apply LPT for neural architecture search in image classification tasks. Following~\citep{liu2018darts}, we first perform architecture search which finds out an optimal cell, then perform architecture evaluation which composes multiple copies of the searched cell into a large network, trains it from scratch, and evaluates the trained model on the test set. We let the target task of the learner and that of the tester be the same.

\subsubsection{Datasets}
\label{sec:datasets}
We used three datasets in the experiments: CIFAR-10, CIFAR-100,  and ImageNet~\citep{deng2009imagenet}. The CIFAR-10 dataset contains 50K training images and 10K testing images, from 10 classes (the number of images in each class is equal). Following~\citep{liu2018darts}, we split the original 50K training set into a new 25K training set and a 25K validation set. In the sequel, when we mention ``training set", it always refers to the new 25K training set. During architecture search, the training set is used as the training data $D_{ee}^{(\textrm{tr})}$ of the learner and the training data $D_{er}^{(\textrm{tr})}$ of the tester. The validation set is used as the test bank $D_b$ and the validation data $D_{er}^{(\textrm{val})}$ of the tester. During architecture evaluation, the combination of the training data and validation data is used to train the large network stacking multiple copies of the searched cell. The CIFAR-100 dataset contains 50K training images and 10K testing images, from 100 classes (the number of images in each class is equal). Similar to CIFAR-100, the 50K training images are split into a 25K training set and 25K validation set. The usage of the new training set and validation set is the same as that for CIFAR-10. The ImageNet dataset contains a training set of 1.2M images and a validation set of  50K images, from 1000 object classes. The  validation set is used as a test set for architecture evaluation. 
Following~\citep{liu2018darts}, we  evaluate the architectures searched using CIFAR-10 and CIFAR-100 on ImageNet: given a cell searched using CIFAR-10 and CIFAR-100, multiple copies of it compose a large network, which is then trained on the 1.2M training data of ImageNet and evaluated on the 50K test data.

\subsubsection{Experimental Settings}
\label{sec:settings}
Our framework is a general one that can be used together with any differentiable search method. Specifically, we apply our framework to the following NAS methods: 1) DARTS~\citep{liu2018darts}, 2) P-DARTS~\citep{chen2019progressive}, 3) DARTS\textsuperscript{+}~\citep{liang2019darts+}, 4) DARTS$^{-}$~\citep{abs-2009-01027}. The search space in these methods are similar. The candidate operations include: $3\times 3$ and $5\times 5$ separable convolutions, $3\times 3$ and $5\times 5$ dilated separable convolutions, $3\times 3$ max pooling, $3\times 3$ average pooling, identity, and zero. In LPT, the network of the learner is a stack of multiple cells, each consisting of 7 nodes. For the data encoder of the tester, we tried ResNet-18 and ResNet-50~\citep{resnet}. For the test creator and target-task executor, they are set to one feed-forward layer. $\lambda$ and $\gamma$ are both set to 1.

For CIFAR-10 and CIFAR-100, during architecture search, the learner's network is a stack of 8 cells, with the initial channel number set to 16. The search is performed for 50 epochs, with a batch size of 64. The hyperparameters for the learner's architecture and weights are set in the same way as DARTS, P-DARTS,  PC-DARTS,  DARTS\textsuperscript{+}, and DARTS$^{-}$. The data encoder and target-task executor of the tester are optimized using SGD with a momentum of 0.9 and a weight decay of 3e-4. The initial learning rate is set to 0.025 with a cosine decay scheduler. The test creator is optimized with the Adam~\citep{adam} optimizer with a learning rate of 3e-4 and a weight decay of 1e-3. During architecture evaluation, 20 copies of the searched cell are stacked to form the learner's network, with the initial channel number set to 36. The network is trained for 600 epochs with a batch size of 96 (for both CIFAR-10 and CIFAR-100). The experiments are performed on a single Tesla v100. For ImageNet, 
following~\citep{liu2018darts}, we take the architecture searched on CIFAR-10 and evaluate it on ImageNet. 
We stack 14 cells (searched on CIFAR-10) to form a large network and set the initial channel number as 48. The network is trained for 250 epochs with a batch size of 1024 on 8 Tesla v100s. Each experiment on LPT is repeated for ten times with the random seed to be from 1 to 10. We report the mean and standard deviation of results obtained from the 10 runs.

\subsubsection{Results}

\begin{table}[t]
    \centering
    \begin{tabular}{l|ccc}
    \toprule
    Method & Error(\%)& Param(M)& Cost\\
    \midrule
    *ResNet \citep{he2016deep}&22.10&1.7&-\\
     *DenseNet \citep{HuangLMW17}&17.18&25.6 &-\\
    \hline
    *PNAS \citep{LiuZNSHLFYHM18}&19.53&3.2&150\\
    *ENAS \citep{pham2018efficient}&19.43&4.6&0.5\\
        *AmoebaNet \citep{real2019regularized}&18.93&3.1&3150\\
    \hline
    *GDAS \citep{DongY19}&18.38&3.4&0.2\\
    *R-DARTS \citep{ZelaESMBH20}&18.01$\pm$0.26&-&1.6
    \\
      *DropNAS \citep{HongL0TWL020} & 16.39&4.4&0.7 \\
\hline
\hline
     ${}^{\dag}$DARTS-1st \citep{liu2018darts}  &20.52$\pm$0.31 &1.8 &0.4\\
     $\;\;$LPT-R18-DARTS-1st (ours) &\textbf{19.11}$\pm$0.11&2.1&0.6 \\
        \hline
            *DARTS-2nd \citep{liu2018darts}  & 20.58$\pm$0.44&1.8&1.5 \\
             $\;\;$LPT-R18-DARTS-2nd (ours) &19.47$\pm$0.20 & 2.1&1.8 \\
               $\;\;$LPT-R50-DARTS-2nd (ours) &\textbf{18.40}$\pm$0.16 &2.5&2.0 \\
            \hline
      *DARTS$^{-}$ \citep{abs-2009-01027}&17.51$\pm$0.25&3.3&0.4\\
      ${}^{\dag}$DARTS$^{-}$ \citep{abs-2009-01027}& 18.97$\pm$0.16& 3.1&0.4\\
           $\;\;$LPT-R18-DARTS$^{-}$ (ours) &\textbf{18.28}$\pm$0.14&3.4& 0.6\\
     \hline
     ${}^{\Delta}$DARTS$^{+}$ \citep{abs-1909-06035}&17.11$\pm$0.43&3.8&0.2\\
                      $\;\;$LPT-R18-DARTS$^{+}$ (ours) &\textbf{16.58}$\pm$0.19& 3.7&0.3 \\
        \hline
      $\dag$PC-DARTS \citep{abs-1907-05737} &17.96$\pm$0.15&3.9&0.1 \\
       $\;\;$LPT-R18-PC-DARTS (ours)&17.04$\pm$0.05&3.6&0.1 \\
         $\;\;$LPT-R50-PC-DARTS (ours)& \textbf{16.97}$\pm$0.21&4.0 &0.1 \\
        \hline
           *P-DARTS \citep{chen2019progressive}&17.49&3.6&0.3\\ 
       $\;\;$LPT-R18-P-DARTS (ours) &\textbf{16.28$\pm$0.10}&3.8& 0.5\\
        $\;\;$LPT-R50-P-DARTS (ours) & 16.38$\pm$0.07& 3.6& 0.5\\
        \bottomrule
    \end{tabular}
    \caption{Results on CIFAR-100, including classification error (\%) on the test set, number of parameters (millions) in the searched architecture, and search cost (GPU days).
    LPT-R18-DARTS-1st denotes that our method LPT is applied to the search space of DARTS. Similar meanings hold for other notations in such a format. R18 and R50 denote that the data encoder of the tester in LPT is set to ResNet-18 and ResNet-50 respectively.  DARTS-1st  and 
DARTS-2nd denotes that first order and second order approximation is used in DARTS.
    * means the results are taken from DARTS$^{-}$ \citep{abs-2009-01027}. $\dag$ means we re-ran this method for 10 times. $\Delta$ means the algorithm ran for 600 epochs instead of 2000 epochs in the architecture evaluation stage, to ensure a fair comparison with other methods (where the epoch number is 600). The search cost is measured by GPU days on a Tesla v100.
    }
    \label{tab:cifar100}
\end{table}

\begin{table}[t]
    \centering
    \begin{tabular}{l|ccc}
    \toprule
    Method& Error(\%)& Param(M) & Cost\\
    \midrule
    *DenseNet
    \citep{HuangLMW17}&3.46&25.6 &-\\
    \hline
     *HierEvol \citep{liu2017hierarchical}&3.75$\pm$0.12& 15.7 &300\\
    *NAONet-WS \citep{LuoTQCL18} & 3.53 & 3.1&0.4 \\
        *PNAS \citep{LiuZNSHLFYHM18} &3.41$\pm$0.09  &3.2& 225\\
        *ENAS \citep{pham2018efficient} &2.89 & 4.6  &0.5 \\
    *NASNet-A \citep{zoph2018learning} & 2.65 & 3.3& 1800\\
    *AmoebaNet-B \citep{real2019regularized} & 2.55$\pm$0.05 & 2.8&3150  \\
    \hline
        *R-DARTS \citep{ZelaESMBH20} &2.95$\pm$0.21  &- & 1.6 \\
            *GDAS \citep{DongY19}&2.93& 3.4& 0.2 \\
    *SNAS \citep{xie2018snas} &2.85 & 2.8& 1.5\\
        *BayesNAS \citep{ZhouYWP19} &2.81$\pm$0.04 &3.4&0.2 \\
        *MergeNAS \citep{WangXYYHS20} &2.73$\pm$0.02 &2.9 & 0.2 \\
        *NoisyDARTS \citep{abs-2005-03566} &2.70$\pm$0.23&3.3  & 0.4 \\
            *ASAP \citep{NoyNRZDFGZ20} &2.68$\pm$0.11 & 2.5&0.2 \\
                *SDARTS
    \citep{abs-2002-05283}&2.61$\pm$0.02 & 3.3& 1.3 \\
            *DropNAS \citep{HongL0TWL020} &2.58$\pm$0.14 & 4.1&0.6 \\
            *PC-DARTS \citep{abs-1907-05737} &2.57$\pm$0.07&3.6& 0.1\\
    *FairDARTS \citep{abs-1911-12126} &2.54 &3.3 &0.4 \\
        *DrNAS \citep{abs-2006-10355} &2.54$\pm$0.03&4.0&  0.4\\
           *P-DARTS \citep{chen2019progressive} &2.50 &3.4&0.3\\
     \hline
        \hline
            *DARTS-1st \citep{liu2018darts} &3.00$\pm$0.14&3.3&  0.4\\
        $\;\;$LPT-R18-DARTS-1st (ours) &\textbf{2.85}$\pm$0.09 &2.7&0.6 \\
        \hline
               *DARTS-2nd \citep{liu2018darts} &2.76$\pm$0.09&3.3&  1.5\\
           $\;\;$LPT-R18-DARTS-2nd (ours)  &2.72$\pm$0.07&3.4& 1.8 \\
            $\;\;$LPT-R50-DARTS-2nd (ours) &  \textbf{2.68}$\pm$0.02  &3.4& 2.0\\
            \hline
            *DARTS$^{-}$ \citep{abs-2009-01027}&2.59$\pm$0.08&  3.5&0.4\\
             ${}^{\dag}$DARTS$^{-}$ \citep{abs-2009-01027}& 2.97$\pm$0.04& 3.3&0.4\\
         $\;\;$LPT-R18-DARTS$^{-}$ (ours) &2.74$\pm$0.07&3.4& 0.6\\
         \hline
             ${}^{\Delta}$DARTS$^{+}$ \citep{abs-1909-06035}&2.83$\pm$0.05&3.7&0.4\\
           $\;\;$LPT-R18-DARTS$^{+}$ (ours) &\textbf{2.69}$\pm$0.05&3.6& 0.5\\
           \hline
                 *PC-DARTS \citep{abs-1907-05737} &\textbf{2.57}$\pm$0.07&3.6& 0.1\\
       $\;\;$LPT-R18-PC-DARTS (ours)& 2.65$\pm$0.17&3.7&0.1\\
     \hline
    *P-DARTS \citep{chen2019progressive}& 2.50&3.4&  0.3\\
     $\;\;$LPT-R18-P-DARTS (ours)& 2.58$\pm$0.14& 3.3 & 0.5 \\
        \bottomrule
    \end{tabular}
    \caption{
    Results on CIFAR-10.
     * means the results are taken from DARTS$^{-}$ \citep{abs-2009-01027}, NoisyDARTS \citep{abs-2005-03566},  and DrNAS \citep{abs-2006-10355}.
    The rest notations are the same as those in Table~\ref{tab:cifar100}.
   }
    \label{tab:cifar10}
\end{table}

\begin{table*}[t]
\small
    \centering
    \begin{tabular}{l|cccc}
    \toprule
  \multirow{2}{*}{Method}   & Top-1  &Top-5 &Param & Cost \\
         & Error (\%) & Error (\%)&(M) & (GPU days)\\
    \midrule
    *Inception-v1 \citep{googlenet}&30.2 &10.1&6.6&- \\
    *MobileNet \citep{HowardZCKWWAA17} &  29.4& 10.5 &4.2&- \\
    *ShuffleNet 2$\times$ (v1) \citep{ZhangZLS18} &  26.4 &10.2 & 5.4&-\\
    *ShuffleNet 2$\times$ (v2) \citep{MaZZS18} &  25.1 &7.6 & 7.4&-\\
    \hline
    *NASNet-A \citep{zoph2018learning} &26.0 &8.4 &5.3 &1800 \\
    *PNAS \citep{LiuZNSHLFYHM18} &25.8 &8.1  &5.1 &225 \\
    *MnasNet-92 \citep{TanCPVSHL19} & 25.2 & 8.0& 4.4&1667\\
        *AmoebaNet-C \citep{real2019regularized} &  24.3 &7.6 &6.4&3150 \\
    \hline
     *SNAS \citep{xie2018snas} & 27.3 &9.2 &4.3 &1.5 \\
          *BayesNAS \citep{ZhouYWP19} &26.5 &8.9 &3.9&0.2 \\
                    *PARSEC \citep{abs-1902-05116} & 26.0 &8.4&5.6&1.0 \\
     *GDAS \citep{DongY19} &  26.0&8.5 &5.3 & 0.2\\
                 *DSNAS \citep{HuXZLSLL20} &25.7& 8.1 &- & -\\
          *SDARTS-ADV \citep{abs-2002-05283}&25.2& 7.8 &5.4& 1.3 \\
           *PC-DARTS \citep{abs-1907-05737} & 25.1 &7.8&5.3&0.1\\
                *ProxylessNAS \citep{cai2018proxylessnas} & 24.9 &7.5 &7.1 &8.3  \\
          *FairDARTS \citep{abs-1911-12126} &24.9 &7.5 &4.8 &0.4 \\
           *P-DARTS (CIFAR-100) \citep{chen2019progressive}&24.7& 7.5&5.1&0.3\\
     *P-DARTS (CIFAR-10) \citep{chen2019progressive}&24.4 &7.4&4.9&0.3\\
     *FairDARTS \citep{abs-1911-12126} &24.4 &7.4 &4.3 &3.0 \\
             *DrNAS \citep{abs-2006-10355} & 24.2 &7.3& 5.2&3.9\\
           *PC-DARTS \citep{abs-1907-05737} &  24.2 &7.3&5.3&3.8\\
         *DARTS$^{+}$ \citep{abs-1909-06035}& 23.9& 7.4&5.1&6.8\\
        *DARTS$^{-}$ \citep{abs-2009-01027}&23.8& 7.0&4.9&4.5\\
     *DARTS$^{+}$ (CIFAR-100) \citep{abs-1909-06035}&23.7& 7.2&5.1&0.2\\
     \hline
      \hline
            *DARTS-2nd-CIFAR-10 \citep{liu2018darts}  & 26.7 &8.7&4.7&4.0 \\
        $\;\;$LPT-R18-DARTS-2nd-CIFAR-10 (ours) & \textbf{25.3}&7.9&4.7&4.0 \\
        \hline
                *P-DARTS (CIFAR10) \citep{chen2019progressive}&24.4 &7.4&4.9&0.3\\
        $\;\;$LPT-R18-P-DARTS-CIFAR10 (ours) & \textbf{24.2}&  7.3&4.9&0.5 \\
        \hline
             *P-DARTS (CIFAR100) \citep{chen2019progressive}&24.7& 7.5&5.1&0.3\\
           $\;\;$LPT-R18-P-DARTS-CIFAR100 (ours) & \textbf{24.0}& 7.1&5.3&0.5\\
           \hline
            *PC-DARTS-ImageNet \citep{abs-1907-05737} &  24.2 &7.3&5.3&3.8\\
               $\;\;$LPT-R18-PC-DARTS-ImageNet (ours)& \textbf{23.4} &  \textbf{6.8}&5.7&4.0\\ 
         \bottomrule
    \end{tabular}
    \caption{Results on ImageNet, including top-1 and top-5 classification errors on the test set, number of weight parameters (millions), and search cost (GPU days). * means the results are taken from DARTS$^{-}$ \citep{abs-2009-01027} and DrNAS \citep{abs-2006-10355}. The rest notations are the same as those in Table~\ref{tab:cifar100}. 
     The first row block shows networks designed by humans manually. 
    The second row block shows non-gradient based search methods.
   The third block shows gradient-based methods. 
    }
    \label{tab:imagenet}
\end{table*}

Table~\ref{tab:cifar100} shows the classification error (\%), number of weight parameters (millions), and search cost (GPU days) of different NAS methods on CIFAR-100. From this table, we make the following observations. \textbf{First}, when our method LPT is applied to different NAS baselines including DARTS-1st (first order approximation), DARTS-2nd (second order approximation), DARTS$^{-}$ (our run), DARTS$^{+}$, PC-DARTS, and P-DARTS, the classification errors of these baselines can be significantly reduced. For example, applying our method to P-DARTS, the error reduces from 17.49\% to 16.28\%. Applying our method to DARTS-2nd, the error reduces from 20.58\% to 18.40\%. 
This demonstrates the effectiveness of our method in searching for a better architecture. In our method, the learner continuously improves its  architecture by passing the tests created by the tester with increasing levels of difficulty. These tests can help the learner to identify the weakness of its architecture and provide guidance on how to improve it. Our method creates a new test on the fly based on how the learner performs in the previous round. From the test bank, the tester selects a subset of difficult examples to evaluate the learner. This new test poses a greater challenge to the learner and encourages the learner to improve its architecture so that it can overcome the new challenge. In contrast, in baseline NAS approaches, a single fixed validation set is used to evaluate the learner. The learner can achieve a good performance via ``cheating": focusing on performing well on the majority of easy examples and ignoring the minority of difficult examples. As a result, the learner's architecture does not have the ability to deal with challenging cases in the unseen data. \textbf{Second}, LPT-R50-DARTS-2nd outperforms LPT-R18-DARTS-2nd, where the former uses ResNet-50 as the data encoder in the tester while the latter uses ResNet-18.  ResNet-50 has a better ability of learning representations than ResNet-18 since it is ``deeper": 50 layers versus 18 layers. 
This shows that a ``stronger" tester can help the learner to learn better. With a more powerful data encoder, the tester can better understand examples in the test bank and can make better decisions in creating difficult and meaningful tests. Tests with better quality can more effectively evaluate the learner and promote its learning capability. \textbf{Third}, our method LPT-R18-P-DARTS  achieves the best performance among all methods, which further demonstrates the effectiveness of LPT in driving the frontiers of neural architecture search forward. \textbf{Fourth}, the number of weight parameters and search costs corresponding to our methods are on par with those in differentiable NAS baselines. This shows that LPT is able to search better-performing architectures without significantly increasing network size and search cost.  A few additional remarks: 1) On CIFAR-100, DARTS-2nd with second-order approximation in the optimization algorithm is not advantageous compared with DARTS-1st which uses first-order approximation; 2) In our run of DARTS$^{-}$, the performance reported in~\citep{abs-2009-01027} cannot be achieved; 3) In our run of DARTS$^+$,  in the architecture evaluation stage, we set the number of epochs to 600  instead of 2000 as used in~\citep{abs-1909-06035},  to ensure a fair comparison with other methods (where the epoch number is 600).

Table~\ref{tab:cifar10} shows the classification error (\%), number of weight parameters (millions), and search cost (GPU days) of different NAS methods on CIFAR-10. As can be seen, applying our proposed LPT to DARTS-1st, DARTS-2nd, DARTS$^{-}$, and DARTS$^{+}$ significantly reduces the errors of these baselines. 
For example, with the usage of LPT, the error of DARTS-2nd is reduced from 2.76\% to 2.68\%.  This further demonstrates the efficacy of our  method in searching better-performing architectures, by creating tests  with increasing levels of difficulty and improving the learner through taking these tests. On PC-DARTS and P-DARTS, applying our method does not yield better performance.

Table~\ref{tab:imagenet} shows the results on ImageNet, including top-1 and top-5 classification errors on the test set. In our proposed LPT-R18-PC-DARTS-ImageNet, the architecture is searched on ImageNet, where our method performs much better than PC-DARTS-ImageNet and achieves the lowest error (23.4\% top-1 error and 6.8\% top-5 error) among all methods in Table~\ref{tab:imagenet}. In our methods including  LPT-R18-P-DARTS-CIFAR100, LPT-R18-P-DARTS-CIFAR10, and LPT-R18-DARTS-2nd-CIFAR10, the architectures are searched on CIFAR-10 or CIFAR-100 and evaluated on ImageNet, where these methods outperform their corresponding baselines P-DARTS-CIFAR100, P-DARTS-CIFAR10,  and DARTS-2nd-CIFAR10. 
These results further demonstrate the effectiveness of our method.

\subsubsection{Ablation Studies}
In order to evaluate the effectiveness of individual modules in LPT, we compare the full LPT framework with the following ablation settings.
\begin{itemize}[leftmargin=*]
    \item \textbf{Ablation setting 1}. In this setting, the tester creates tests solely by maximizing their level of difficulty, without considering their meaningfulness. Accordingly, the second stage in LPT where the tester learns to perform a target-task by leveraging the created tests is removed. 
    The tester 
    directly learns a selection scalar $s(d)\in[0,1]$ for each  example $d$ in the test bank  without going through a data encoder or a test creator. The corresponding formulation is:
        \begin{equation}
    \begin{array}{l}
\max _{S} \min _{A} \;\; \frac{1}{\sum_{d\in D_{b}}s(d)} \sum_{d\in D_{b}} s(d) \ell (A, W^{*}(A),d)\\
s.t. \;\;    W^{*}(A)=\min _{W} \;\; L\left(A, W, D_{ee}^{(\mathrm{tr})}\right)
\end{array}
\end{equation}
where $S=\{s(d)|d\in D_{b}\}$. In this study, $\lambda$ and $\gamma$ are both set to 1. The data encoder of the tester is ResNet-18. For CIFAR-100, to avoid performance collapse because of skip connections, LPT is applied to P-DARTS. For CIFAR-10, LPT is applied to DARTS-2nd.

  \item \textbf{Ablation setting 2}. In this setting, in the second stage of LPT, the tester is trained solely based on the create test, without using the training data of the target task. 
  The corresponding formulation is: 
    \begin{equation}
    \begin{array}{l}
\max _{C} \min _{A} \;\; \frac{1}{\left|\sigma\left(C, E^{*}(C), D_{b}\right)\right|} L\left(A, W^{*}\left(A\right), \sigma\left(C, E^{*}(C), D_{b}\right)\right)-\lambda L\left(E^{*}(C), X^{*}(C), D_{er}^{(\mathrm{val})}\right) \\
s.t. \;\;   E^{*}(C), X^{*}(C)=\min _{E, X} \;\; L\left(E, X, \sigma\left(C, E, D_{b}\right)\right) \\
\quad\;\;\; W^{*}\left(A\right)=\min _{W}\;\;L\left(A, W, D_{ee}^{(\mathrm{tr})}\right)
\end{array}
\end{equation}
In this study,  $\lambda$ and $\gamma$ are both set to 1. The data encoder of the tester is ResNet-18. For CIFAR-100, to avoid performance collapse because of skip connections, LPT is applied to P-DARTS. For CIFAR-10, LPT is applied to DARTS-2nd.
    \item Ablation study on $\lambda$. We are interested in how the learner's performance varies as the tradeoff parameter $\lambda$ in Eq.(\ref{eq:learning_objective}) increases. In this study, the other tradeoff parameter $\gamma$ in Eq.(\ref{eq:learning_objective}) is set to 1. For both CIFAR-100 and CIFAR-10, we randomly sample 5K data from the 25K training  and  25K validation data, and use it as a test set to report performance in this ablation study. 
    The rest 45K data (22.5K training data and 22.5K validation data) is used for architecture search and evaluation. Tester's data encoder is ResNe-18. LPT is applied to P-DARTS.
    \item Ablation study on $\gamma$. We investigate how the learner's performance varies as $\gamma$ increases. In this study, the other tradeoff parameter $\lambda$ is set to 1. Similar to the ablation study on $\lambda$, on 5K randomly-sampled test data, we report performance of architectures searched and evaluated on 45K data. Tester's data encoder is ResNe-18. LPT is applied to P-DARTS.
 \end{itemize}

\begin{table}[t]
    \centering
    \begin{tabular}{l|c}
    \hline
    Method & Error (\%)\\
    \hline
    Difficulty only (CIFAR-100) &  18.12$\pm$0.11 \\ 
            Difficulty + meaningfulness (CIFAR-100) &\textbf{17.18}$\pm$0.12  \\
         \hline
             Difficulty only (CIFAR-10) & 2.79$\pm$0.06  \\
         Difficulty + meaningfulness (CIFAR-10) &\textbf{2.72}$\pm$0.07   \\
         \hline
    \end{tabular}
    \caption{Results for ablation setting 1. ``Difficulty only" denotes that the tester creates tests solely by maximizing their level of difficulty,  without considering their meaningfulness, i.e., the tester does not use the tests to learn to perform the target task. ``Difficulty + meaningfulness" denotes the full LPT framework where the tester creates tests by maximizing both  difficulty and meaningfulness. }
    \label{tab:ab1}
\end{table}

\begin{table}[t]
    \centering
    \begin{tabular}{l|c}
    \hline
    Method & Error (\%)\\
    \hline
         Test only (CIFAR-100) & 17.54$\pm$0.07 \\ 
           Test + Training data (CIFAR-100) &\textbf{17.18}$\pm$0.12  \\
         \hline
          Test only (CIFAR-10) & 2.75$\pm$0.03  \\
         Test + Training data (CIFAR-10) &\textbf{2.72}$\pm$0.07  \\
         \hline
    \end{tabular}
    \caption{
    Results for ablation setting 2. ``Test only" denotes that the tester is trained only using the create test to perform the target task. ``Test + Training data" denotes that the tester is trained using both the test and the training data of the target task. 
   }
    \label{tab:ab4}
\end{table}

Table~\ref{tab:ab1} shows the results for ablation setting 1. As can be seen, on both CIFAR-10 and CIFAR-100, creating tests that are both difficult and   meaningful is better than creating tests solely by maximizing difficulty. The reason is that a difficult test could be composed of bad-quality examples such as outliers and incorrectly-labeled examples. Even a highly-accurate learner model cannot achieve good performance on such erratic examples. To address this problem, it is necessary to make the created tests meaningful. LPT achieves meaningfulness of the tests by making the tester leverage the created tests to perform the target task. The results demonstrate that this is an effective way of improving meaningfulness.

Table~\ref{tab:ab4} shows the results for ablation setting 2. As can be seen, for both CIFAR-100 and CIFAR-10, using both the created test and the training data of the target task to train the tester performs better than using the test only. By leveraging the training data, the data encoder can be better trained. And a better encoder can help to create higher-quality tests.

\begin{figure}[H]
    \centering
 \includegraphics[width=0.49\columnwidth]{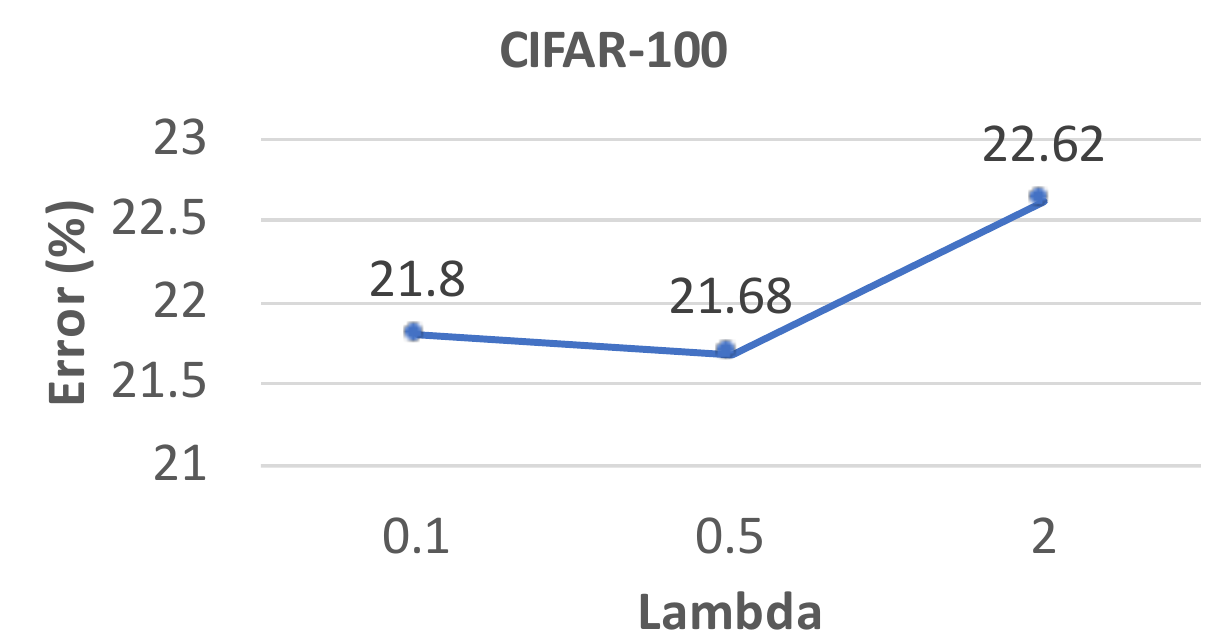}
  \includegraphics[width=0.49\columnwidth]{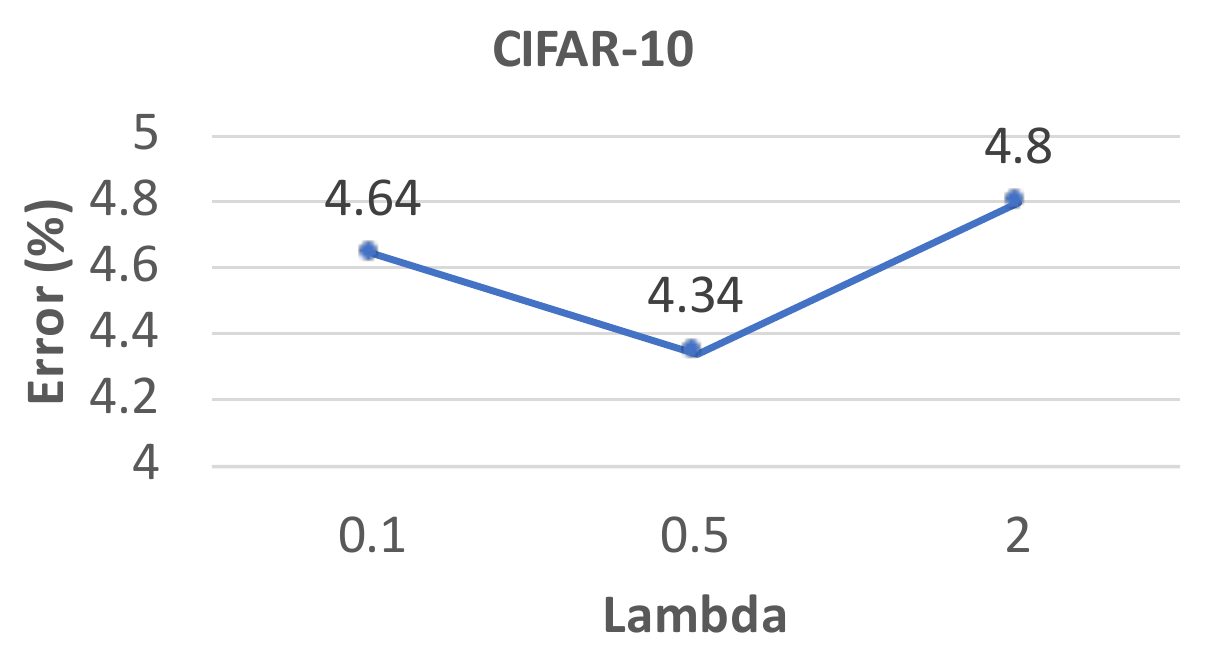}
       \caption{How errors change as $\lambda$ increases.}
 \label{fig:lambda}
\end{figure}

\begin{figure}[H]
    \centering
 \includegraphics[width=0.49\columnwidth]{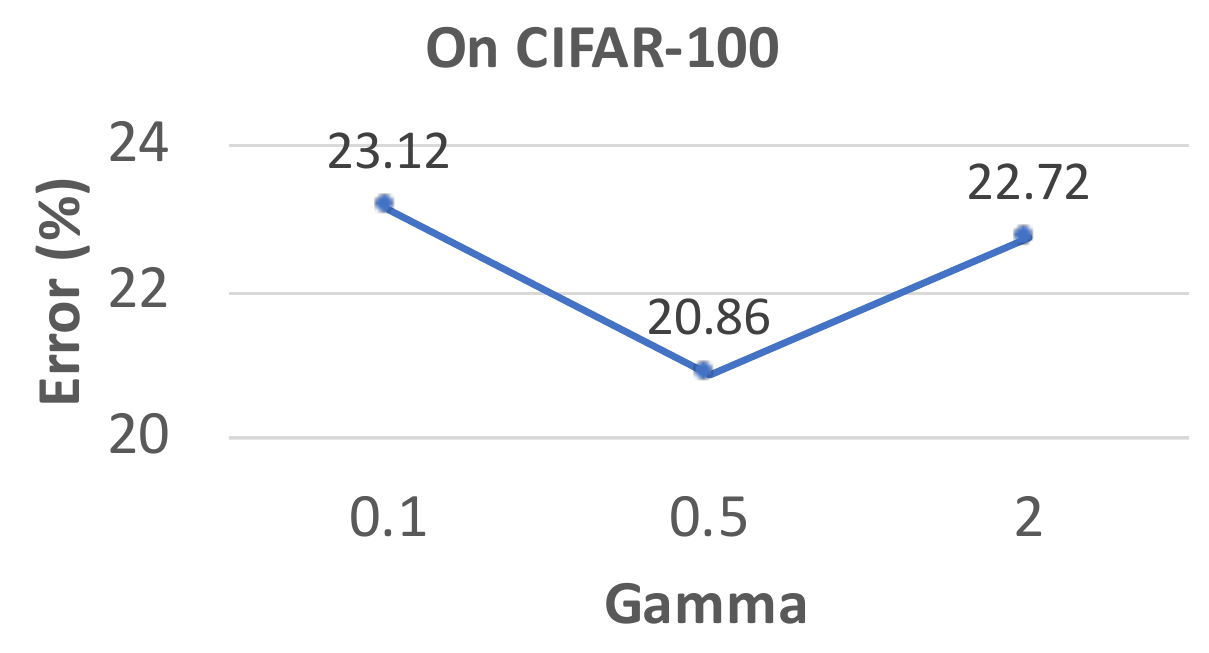}
  \includegraphics[width=0.49\columnwidth]{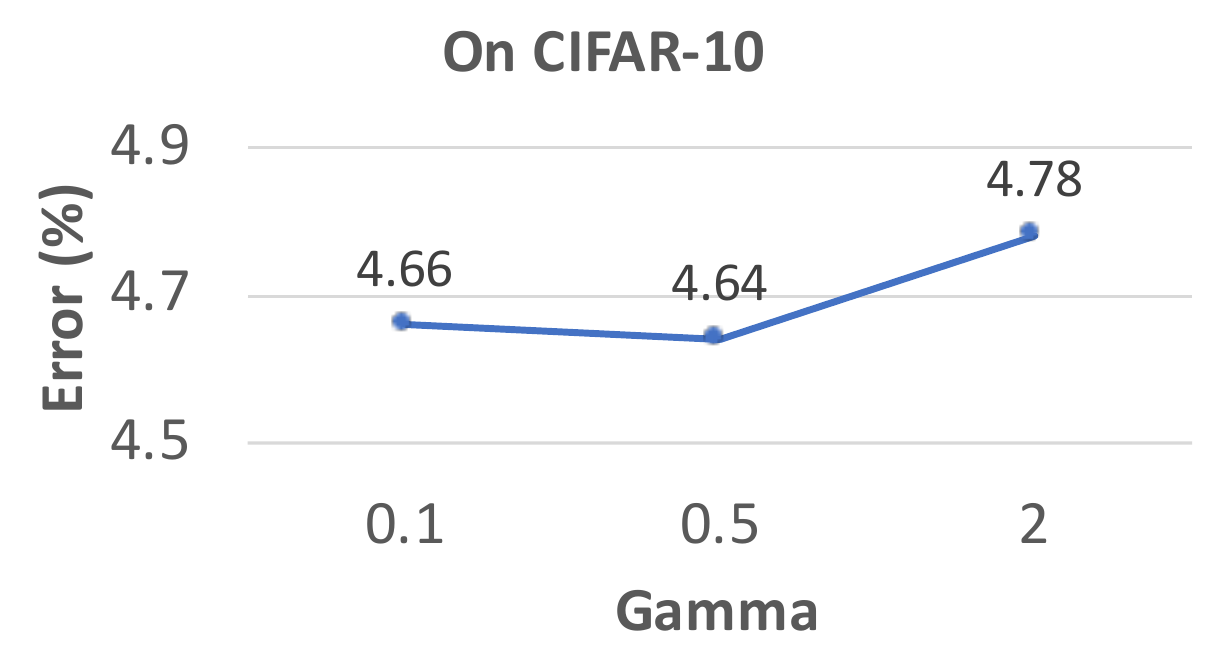}
       \caption{How errors change as $\gamma$ increases.}
 \label{fig:gamma}
\end{figure}

Figure~\ref{fig:lambda} shows how classification errors change as $\lambda$ increases. As can be seen, on both CIFAR-100 and CIFAR-10, when $\lambda$ increases from 0.1 to 0.5, the error decreases. However, further increasing $\lambda$ renders the error to increase. From the tester's perspective, $\lambda$ explores a tradeoff between difficulty and meaningfulness of the tests. Increasing $\lambda$ encourages the tester to create tests that are more meaningful.  Tests with more meaningfulness can more reliably evaluate the learner. However, if $\lambda$ is too large, the tests are biased to be more meaningful and less difficult. Lacking enough difficulty, the tests may not be compelling enough to drive the learner for improvement. Such a tradeoff effect is observed in the results on CIFAR-10 as well.

Figure~\ref{fig:gamma} shows how classification errors change as $\gamma$ increases. As can be seen, on both CIFAR-100 and CIFAR-10, when $\gamma$ increases from 0.1 to 0.5, the error decreases. However, further increasing $\gamma$ renders the error to increase. Under a larger  $\gamma$, the created test plays a larger role in training the tester to perform the target task. This implicitly encourages the test creator to generate tests that are more meaningful. However, if $\gamma$ is too large, the training is dominated by the created test which incurs the following risk: if the test is not meaningful, it will result in a poor-quality data-encoder which further degrades the quality of test creation.

\subsection{Summary}
In this section, we apply Skillearn to formalize a skill in human learning -- learning by passing tests (LPT) and use it for neural architecture search. 
In LPT, a tester model creates a sequence of tests with growing levels of difficulty. A learner model continuously improves its learning ability by striving to pass these increasingly more-challenging tests. 
 The tester learns to  select hard validation examples  rendering the learner to make large prediction errors and the learner refines its model to rectify these prediction errors. Our framework  achieves significant improvement in neural architecture search on  CIFAR-100, CIFAR-10, and ImageNet. 

\section{Case Study II: Interleaving Learning}
In this section, we instantiate our general Skillearn framework to formalize another human learning
technique – interleaving learning, and apply it to improve machine learning. Interleaving learning is a learning technique where a learner interleaves the studies of multiple topics: study topic $A$ for a while, then switch to $B$, subsequently to $C$; then switch back to $A$, and so on, forming a pattern of $ABCABCABC\cdots$. Interleaving learning is in contrast to blocked learning, which studies one topic very thoroughly before moving to another topic. Compared with blocked learning, interleaving learning increases long-term retention and improves ability to transfer learned knowledge. 

We are interested in investigating whether the interleaving strategy is helpful for training machine learning models. We instantiate the Skillearn framework to an interleaving learning (IL) framework. We assume there are $K$ learning tasks, each performed by a learner model. Each learner has a data encoder and a task-specific head. The data encoders of all learners share the same architecture, but may have different weight parameters. The $K$ learners perform $M$ rounds of interleaving learning with the following order: 
\begin{equation}
\underbrace{l_1,l_2,\cdots,l_K}_{\textrm{Round }1}
\underbrace{l_1,l_2,\cdots,l_K}_{\textrm{Round }2}
\cdots
\underbrace{l_1,l_2,\cdots,l_K}_{\textrm{Round }m}
\cdots
\underbrace{l_1,l_2,\cdots,l_K}_{\textrm{Round }M}
\end{equation}
where $l_k$ denotes that the $k$-th learner performs learning. In the first round, we first learn $l_1$, then learn $l_2$, and so on. At the end of the first round, $l_K$ is learned. Then we move to the second round, which starts with learning $l_1$, then learns $l_2$, and so on. This pattern repeats until the $M$ rounds of learning are finished. Between two consecutive learners $l_kl_{k+1}$, the encoder weights of the latter learner $l_{k+1}$ are encouraged to be close to the optimally learned encoder weights of the former learner $l_k$.

\subsection{Method}
\begin{table}[t]
\centering
\begin{tabular}{l|p{12cm}}
\hline
Notation & Meaning \\
$K$ & Number of learners\\
$M$ & Number of rounds\\
$D_k^{(\textrm{tr})}$ & Training dataset of the $k$-th learner\\
$D_k^{(\textrm{val})}$ & Validation dataset of the $k$-th learner\\
$A$ & Encoder architecture shared by all learners\\
$W_k^{(m)}$ & Weight parameters in the data encoder of the $k$-th learner in the $m$-th round\\
$H_k^{(m)}$ & Weight parameters in the task-specific head of the $k$-th learner in the $m$-th round\\
$\widetilde{W}_k^{(m)}$ & The optimal  encoder weights of the $k$-th learner in the $m$-th round\\
$\widetilde{H}_k^{(m)}$ & The optimal weight parameters of the task-specific  head  in the $k$-th learner in the $m$-th round\\
$\gamma$ & Tradeoff parameter\\
\hline
\end{tabular}
\caption{Notations in interleaving learning}
\label{tb:notations}
\end{table}

In this section, we present the details of the interleaving learning framework. There are $K$ learners. Each learner learns to perform a task. These tasks could be the same, e.g., image classification on CIFAR-10; or different, e.g., image classification on CIFAR-10, image classification on ImageNet~\citep{deng2009imagenet}, object detection on MS-COCO~\citep{coco}, etc. Each learner $k$ has a training dataset $D_k^{(\textrm{tr})}$ and a validation dataset $D_k^{(\textrm{val})}$. 
Each learner has a data encoder and a task-specific head performing the target task. For example, if the task is image classification, the data encoder could be a convolutional neural network extracting visual features of the input images and the task-specific head could be a multi-layer perceptron which takes the visual features of an image extracted by the data encoder as input and predicts the class label of this image. We assume the architecture of the data encoder in each learner is learnable. The data encoders of all learners share the same architecture, but their weight parameters could be different in different learners. The architectures of task-specific heads are manually designed by humans  and they could be different in different learners. The $K$ learners perform $M$ rounds of interleaving learning with the following order: 
\begin{equation}
\underbrace{l_1,l_2,\cdots,l_K}_{\textrm{Round }1}
\underbrace{l_1,l_2,\cdots,l_K}_{\textrm{Round }2}
\cdots
\underbrace{l_1,l_2,\cdots,l_K}_{\textrm{Round }m}
\cdots
\underbrace{l_1,l_2,\cdots,l_K}_{\textrm{Round }M}
\end{equation}
where $l_k$ denotes that the $k$-th learner performs learning. In the first round, we first learn $l_1$, then learn $l_2$, and so on. At the end of the first round, $l_K$ is learned. Then we move to the second round, which starts with learning $l_1$, then learns $l_2$, and so on. This pattern repeats until the $M$ rounds of learning are finished. Between two consecutive learners $l_kl_{k+1}$, the weight parameters of the latter learner $l_{k+1}$ are encouraged to be close to the optimally learned encoder weights of the former learner $l_k$. For each learner, the architecture of its encoder remains the same across all rounds; the weights of the encoder and head can be different in different rounds. 

Each learner $k$ has the following learnable parameter sets: 1) architecture $A$ of the encoder; 2) in each round $m$, the learner's encoder has a set of weight parameters $W_k^{(m)}$ specific to this round; 3) in each round $m$, the learner's task-specific head  has a set of weight parameters $H_k^{(m)}$ specific to this round. The encoders of all learners share the same architecture and this architecture remains the same in different rounds. The encoders of different learners have different weight parameters. The weight parameters of a learner's encoder are different in different rounds. Different learners have different task-specific  heads in terms of both architectures and weight parameters. In the interleaving process, the learning of the $k$-th learner is assisted by  the $(k-1)$-th learner. Specifically, during learning, the encoder weights $W_{k}$ of the $k$-th learner are encouraged to be close to the optimal encoder weights $\widetilde{W}_{k-1}$ of the  $(k-1)$-th learner. This is achieved by minimizing an interactive function: $\|W_{k}-\widetilde{W}_{k-1}\|_2^2$.

\begin{table}[t]
\centering
\begin{tabular}{l|p{8cm}}
\hline
Active learners & The first learner\\
\hline
Active learnable parameters & Weights of the data encoder and weights of the task-specific head in the first learner\\
\hline
Supporting learnable parameters & Encoder architecture shared by all learners \\
\hline
Active training datasets &  Training dataset of the first learner\\
\hline
Active auxiliary datasets & --  \\
\hline
Training loss &  The first learner trains the weights of its data encoder and the weights of its task-specific head on its training dataset: $L(A, W^{(1)}_1, H^{(1)}_1,D_1^{(\textrm{tr})})$.  \\
\hline
Interaction function & -- \\
\hline
Optimization problem &  $\widetilde{W}_1^{(1)}(A) =\textrm{min}_{W^{(1)}_1,H^{(1)}_1} \; L(A, W^{(1)}_1, H^{(1)}_1,D_1^{(\textrm{tr})})$ \\
\hline
\end{tabular}
\caption{Learning stage 1 in interleaving learning}
\label{tb:il-1}
\end{table}

There are $M\times K$ learning stages: in each of the $M$ rounds, each of the $K$ learners is learned in a stage. In the very first learning stage, the first learner in the first round is learned. It trains the weight parameters of its data encoder and the weight parameters of its task-specific head on its training dataset. In this learning stage (Table~\ref{tb:il-1}), the active learner is the first learner. The active learnable parameters are the weight parameters of the data encoder and the weight parameters of the task-specific head in the first learner in the first round. The supporting learnable parameters include the encoder architecture shared by all learners. The active training dataset is the training data of the first learner. There is no auxiliary dataset. The training loss is the target-task's loss defined on the training dataset of the first learner: $L(A, W^{(1)}_1, H^{(1)}_1,D_1^{(\textrm{tr})})$. There is no interaction function. The optimization problem is:
\begin{equation}
\widetilde{W}_1^{(1)}(A) =\textrm{min}_{W^{(1)}_1,H^{(1)}_1} \; L(A, W^{(1)}_1, H^{(1)}_1,D_1^{(\textrm{tr})})
\end{equation}
In this optimization problem, $A$ is not learned. After learning, the optimal head is discarded. The optimal encoder weights $\widetilde{W}_1^{(1)}(A)$ are a function of $A$ since the training loss is a function of $A$ and $\widetilde{W}_1$ is a function of the training loss.  $\widetilde{W}_1^{(1)}(A)$ is passed to the next learning stage to help with the learning of the second learner. 

In any other learning stage (Table~\ref{tb:il-k}), e.g., the $l$-th stage where the learner is $k$ and the round of interleaving is $m$, the active learner is the learner $k$. The active learnable parameters include weights of the data encoder and weights of the task-specific head in the $k$-th learner in the $m$-th round. The supporting learnable parameters are the encoder architecture shared by all learners. The active training dataset is the training dataset of the $k$-th learner. There is no active auxiliary dataset. The training loss is the target-task's loss defined on the training dataset of the $k$-th learner: $L(A, W^{(m)}_k, H^{(m)}_k,D_k^{(\textrm{tr})})$. The interaction function $\|W^{(m)}_k-\widetilde{W}_{l-1}\|_2^2$ encourages the encoder weights $W^{(m)}_k$ at this stage to be close to the optimal encoder weights $\widetilde{W}_{l-1}$ learned in the previous stage. The optimization problem is:
\begin{equation}
   \widetilde{W}_k^{(m)}= \textrm{min}_{W_k^{(m)},H_k^{(m)}} \; L(A, W_k^{(m)}, H_k^{(m)},D_k^{(\textrm{tr})})+\lambda\|W^{(m)}_k-\widetilde{W}_{l-1}(A)\|^2_{2}
\end{equation}
where $\lambda$ is a tradeoff parameter. 

The optimal encoder weights are a function of the encoder architecture. The encoder architecture is not updated at this learning stage. In the round of 1 to $M-1$, the optimal heads are discarded after learning. In the round of $M$, the optimal heads are retained and will be used in the validation stage. 

\begin{table}[t]
\centering
\begin{tabular}{l|p{8cm}}
\hline
Active learners & The $k$-th learner \\
\hline
Active learnable parameters & Weights of the data encoder and weights of the task-specific head in  the $k$-th learner\\
\hline
Supporting learnable parameters & Encoder architecture shared by all learners\\
\hline
Active training datasets & Training dataset of the $k$-th learner \\
\hline
Active auxiliary datasets & --  \\
\hline
Training loss &  The $k$-th learner trains the weights of its data encoder and the weights of its task-specific head on its training dataset: $L(A, W^{(m)}_k, H^{(m)}_k,D_k^{(\textrm{tr})})$ \\
\hline
Interaction function & The learner encourages its  encoder weights to be close to the optimal encoder weights  $\widetilde{W}_{l-1}$ learned in the $l-1$ stage: $\|W^{(m)}_k-\widetilde{W}_{l-1}\|_2^2$ \\
\hline
Optimization problem & $\widetilde{W}_k^{(m)}=\textrm{min}_{W_k^{(m)},H_k^{(m)}} \; L(A, W_k^{(m)}, H_k^{(m)},D_k^{(\textrm{tr})})+\lambda\|W^{(m)}_k-\widetilde{W}_{l-1}(A)\|^2_{2}$ \\
\hline
\end{tabular}
\caption{Learning stage $l$ with the $k$-th learner at the $m$-th round, in interleaving learning}
\label{tb:il-k}
\end{table}

\begin{table}[t]
\centering
\begin{tabular}{l|p{8cm}}
\hline
Active learners & All learners\\
\hline
Remaining learnable parameters & Encoder's architecture of all learners\\
\hline
Validation datasets& Validation datasets of all learners\\
\hline
Active auxiliary datasets & -- \\
\hline
Validation loss & The sum of every learner's validation loss on its validation dataset: $\sum_{k=1}^K L(A, \widetilde{W}_k^{(M)}(A), \widetilde{H}_k^{(M)}(A),D_k^{(\textrm{val})}) $ \\
\hline
Interaction function & -- \\
\hline
Optimization problem & $\textrm{min}_{A} \;
  \sum_{k=1}^K L(A, \widetilde{W}_k^{(M)}(A), \widetilde{H}_k^{(M)}(A),D_k^{(\textrm{val})})  $ \\
\hline
\end{tabular}
\caption{Validation stage in interleaving learning}
\label{tb:il-val}
\end{table}

In the validation stage (Table~\ref{tb:il-val}), the active learners are all $K$ learners. The remaining learnable parameters are the encoder architecture shared by all learners. The validation datasets are the validation datasets of all learners. 
There is no active auxiliary dataset. The validation loss is the sum of every learner's validation loss calculated using the optimal encoder weights and head weights learned in the final round: $\sum_{k=1}^K L(A, \widetilde{W}_k^{(M)}(A), \widetilde{H}_k^{(M)}(A),D_k^{(\textrm{val})})$.  There is no interaction function. The optimization problem is:
\begin{equation}
    \textrm{min}_{A} \;
  \sum_{k=1}^K L(A, \widetilde{W}_k^{(M)}(A), \widetilde{H}_k^{(M)}(A),D_k^{(\textrm{val})}).  
\end{equation}

Putting all these pieces together, we instantiate the Skillearn framework to an interleaving learning framework, as shown in Eq.(\ref{eq:il}). 
From bottom to top, the $K$ learners perform $M$ rounds of interleaving learning. Learners in adjacent learning stages are coupled via the interaction function. The architecture $A$ is not updated in the learning stages. It is learned by minimizing the validation loss. Table~\ref{tb:sk-il} summarizes the key elements of interleaving learning under the Skillearn terminology.

\begin{equation}
\begin{array}{ll}
   \textrm{min}_{A}  & 
  \sum_{k=1}^K L(A, \widetilde{W}_k^{(M)}(A), \widetilde{H}_k^{(M)}(A),D_k^{(\textrm{val})}) 
   \\
      s.t.  
        & \textrm{\textbf{Round $\mathbf{M}$}:}\\
      & \widetilde{W}_K^{(M)}(A),\widetilde{H}_K^{(M)}(A) =\textrm{min}_{W_K^{(M)},H^{(M)}_K} \quad L(A, W_K^{(M)}, H^{(M)}_K,D_K^{(\textrm{tr})})+\lambda\|W^{(M)}_K-\widetilde{W}_{K-1}^{(M)}(A)\|^2_{2}\\
      & \cdots \\
    & \widetilde{W}_1^{(M)}(A), \widetilde{H}_1^{(M)}(A) =\textrm{min}_{W^{(M)}_1,H^{(M)}_1} \quad L(A, W^{(M)}_1, H^{(M)}_1,D_1^{(\textrm{tr})})+\lambda\|W^{(M)}_1-\widetilde{W}_K^{(M-1)}(A)\|^2_{2}\\
    & \cdots\\
        & \textrm{\textbf{Round 2}:}\\
      & \widetilde{W}_K^{(2)}(A) =\textrm{min}_{W_K^{(2)},H^{(2)}_K} \quad L(A, W_K^{(2)}, H^{(2)}_K,D_K^{(\textrm{tr})})+\lambda\|W^{(2)}_K-\widetilde{W}_{K-1}^{(2)}(A)\|^2_{2}\\
      & \cdots \\
    & \widetilde{W}_1^{(2)}(A) =\textrm{min}_{W^{(2)}_1,H^{(2)}_1} \quad L(A, W^{(2)}_1, H^{(2)}_1,D_1^{(\textrm{tr})})+\lambda\|W^{(2)}_1-\widetilde{W}_K^{(1)}(A)\|^2_{2}\\
   & \textrm{\textbf{Round 1}:}\\
      & \widetilde{W}_K^{(1)}(A) =\textrm{min}_{W_K^{(1)},H^{(1)}_K} \quad L(A, W_K^{(1)}, H^{(1)}_K,D_K^{(\textrm{tr})})+\lambda\|W^{(1)}_K-\widetilde{W}_{K-1}^{(1)}(A)\|^2_{2}\\
      & \cdots \\
      & \widetilde{W}_k^{(1)}(A) =\textrm{min}_{W_k^{(1)},H^{(1)}_k} \quad L(A, W_k^{(1)}, H^{(1)}_k,D_k^{(\textrm{tr})})+\lambda\|W^{(1)}_k-\widetilde{W}_{k-1}^{(1)}(A)\|^2_{2}\\
      & \cdots \\
      & \widetilde{W}_2^{(1)}(A) =\textrm{min}_{W_2^{(1)},H^{(1)}_2} \quad L(A, W^{(1)}_2, H^{(1)}_2,D_2^{(\textrm{tr})})+\lambda\|W^{(1)}_2-\widetilde{W}_1^{(1)}(A)\|^2_{2}\\
    & \widetilde{W}_1^{(1)}(A) =\textrm{min}_{W^{(1)}_1,H^{(1)}_1} \quad L(A, W^{(1)}_1, H^{(1)}_1,D_1^{(\textrm{tr})})
\end{array}
\label{eq:il}
\end{equation}

\begin{table}[t]
\centering
\begin{tabular}{l|p{10.5cm}}
\hline
Skillearn & Interleaving Learning\\
\hline 
Learners & $K$ learners \\
\hline
Learnable parameters &  1) Encoder architecture shared by all learners; 2) In each round, each learner has weight parameters for the data encoder and weight parameters for the task-specific head.  \\
\hline
Interaction function &  The encoder weights $W_l$ at learning stage $l$ are encouraged to be close to the optimal  encoder weights $\widetilde{W}_{l-1}$ at stage $l-1$: $\|W_l-\widetilde{W}_{l-1}\|_2^2$.\\
\hline
Learning stages &  1) In the first learning stage (the first learner in the first round), the learner trains the weights of its data  encoder  and the weights of its task-specific head on its training dataset: $\widetilde{W}_1^{(1)}(A) =\textrm{min}_{W^{(1)}_1,H^{(1)}_1} \quad L(A, W^{(1)}_1, H^{(1)}_1,D_1^{(\textrm{tr})})$; 2) In other learning stages, the learner trains the weights of its data encoder and the weights of its task-specific head on its training dataset where the  encoder weights are encouraged to be close to the optimal  encoder weights trained in the previous stage: $\widetilde{W}_k^{(m)}(A) =\textrm{min}_{W_k^{(m)},H^{(m)}_k} \quad L(A, W^{(m)}_k, H^{(m)}_k,D_k^{(\textrm{tr})})+\lambda\|W^{(m)}_k-\widetilde{W}_{k-1}^{(m)}(A)\|^2_{2}$. 
\\
\hline 
Validation stage & Each learner validates its optimal data encoder and task-specific head learned in the last round on its validation dataset.\\
\hline 
Datasets & Each learner has a training dataset and a validation dataset.\\
\hline
\end{tabular}
\caption{Mapping from Skillearn to Interleaving Learning}
\label{tb:sk-il}
\end{table}

\subsection{Optimization Algorithm}
In this section,  we develop an optimization algorithm for interleaving learning by instantiating the general optimization framework of Skillearn in Section~\ref{opt:sk}. For each optimization problem $\widetilde{W}_k^{(m)}(A) =\textrm{min}_{W^{(m)}_k,H^{(m)}_k} \quad L(A, W^{(m)}_k, H^{(m)}_k,D_k^{(\textrm{tr})})+\lambda\|W^{(m)}_k-\widetilde{W}_{k-1}^{(m)}(A)\|^2_{2}
$ in a learning stage, we approximate the optimal solution $\widetilde{W}_k^{(m)}(A)$ by one-step gradient descent update of the optimization variable $W^{(m)}_k$: 
\begin{equation}
    \widetilde{W}_k^{(m)}(A)\approx \overline{W}_k^{(m)}(A)= W^{(m)}_k-\eta \nabla_{W^{(m)}_k}( L(A, W^{(m)}_k, H^{(m)}_k,D_k^{(\textrm{tr})})+\lambda\|W^{(m)}_k-\widetilde{W}_{k-1}^{(m)}(A)\|^2_{2})
\end{equation}
For $\widetilde{W}_1^{(1)}(A)$, the approximation is:
\begin{equation}
    \widetilde{W}_1^{(1)}(A)\approx  \overline{W}_1^{(1)}(A)= W^{(1)}_1-\eta \nabla_{W^{(1)}_1} L(A, W^{(1)}_1, H^{(1)}_1,D_1^{(\textrm{tr})})
    \label{eq:w11}
\end{equation}
For $\widetilde{W}_k^{(m)}(A)$, the approximation is:
\begin{equation}
    \widetilde{W}_k^{(m)}(A)\approx \overline{W}_k^{(m)}(A)= W^{(m)}_k-\eta \nabla_{W^{(m)}_k} L(A, W^{(m)}_k, H^{(m)}_k,D_k^{(\textrm{tr})})-2\eta\lambda(W^{(m)}_k-\overline{W}_{k-1}^{(m)}(A))
    \label{eq:wkm}
\end{equation}
where $\overline{W}_{k-1}^{(m)}(A)$ is the approximation of $\widetilde{W}_{k-1}^{(m)}(A)$. Note that $\{\overline{W}_k^{(m)}(A)\}_{k,m=1}^{K,M}$ are  calculated recursively, where $\overline{W}_k^{(m)}(A)$ is a function of $\overline{W}_{k-1}^{(m)}(A)$, $\overline{W}_{k-1}^{(m)}(A)$ is a function of $\overline{W}_{k-2}^{(m)}(A)$, and so on. When $m>1$ and $k=1$,  $\overline{W}_{k-1}^{(m)}(A)=\overline{W}_{K}^{(m-1)}(A)$. For $\widetilde{H}_k^{(M)}(A)$, the approximation is:
\begin{equation}
    \widetilde{H}_k^{(M)}(A)\approx \overline{H}_k^{(M)}(A)=H_k^{(M)}(A)-\eta \nabla_{H_k^{(M)}(A)}L(A, W_k^{(M)}, H^{(M)}_k,D_k^{(\textrm{tr})})
    \label{eq:hm}
\end{equation}
In the validation stage, we plug in the approximations of $\{\widetilde{W}_k^{(M)}(A)\}_{k=1}^K$ and $\{\widetilde{H}_k^{(M)}(A)\}_{k=1}^K$ into the validation loss function, calculate the gradient of the approximated objective w.r.t the encoder architecture $A$, then update $A$ via:
\begin{equation}
    A\gets A-\eta \sum_{k=1}^K \nabla_AL(A, \overline{W}_k^{(M)}(A), \overline{H}_k^{(M)}(A),D_k^{(\textrm{val})}) 
    \label{eq:a-il}
\end{equation}
The update steps from Eq.(\ref{eq:w11}) to Eq.(\ref{eq:a-il}) until convergence. The entire algorithm is summarized in Algorithm~\ref{algo:algo-il}. 

\begin{algorithm}[H]
\SetAlgoLined
 \While{not converged}{
1. Update $\widetilde{W}_1^{(1)}(A)$ using Eq.(\ref{eq:w11})\\
2. For $k=2\cdots K$, update $\widetilde{W}_k^{(1)}(A)$ using Eq.(\ref{eq:wkm})\\
3. For $k=1\cdots K$ and $m=2\cdots M$, update $\widetilde{W}_k^{(m)}(A)$ using Eq.(\ref{eq:wkm})\\
4. For $k=1\cdots K$, update $\widetilde{H}_k^{(M)}(A)$ using Eq.(\ref{eq:hm})\\
5. Update $A$ using Eq.(\ref{eq:a-il})
 }
 \caption{Optimization algorithm for interleaving learning}
 \label{algo:algo-il}
\end{algorithm}

\subsection{Experiments}
We apply interleaving learning for neural architecture search in image classification tasks. Two tasks are interleaved: image classification on CIFAR-10 and image classification on CIFAR-100. We search the shared architecture of data encoders in these two tasks.

\begin{table}[t]
\caption{Classification error (\%) on the test set of CIFAR-100, number of parameters (millions) in the searched architecture, and search cost (GPU days). DARTS-1st and DARTS-2nd denotes that first-order and second-order approximations are used in DARTS.
    * denotes that the results are taken from DARTS$^{-}$ \citep{abs-2009-01027}.  $\dag$ denotes that this approach was re-run  for 10 times.
    The search cost is measured by GPU days on a Tesla v100.
    }
    \centering
    \begin{tabular}{l|ccc}
    \toprule
    Method & Error(\%)& Param(M)& Cost\\
    \midrule
    *ResNet \citep{he2016deep}&22.10&1.7&-\\
     *DenseNet \citep{HuangLMW17}&17.18&25.6 &-\\
    \hline
    *PNAS \citep{LiuZNSHLFYHM18}&19.53&3.2&150\\
    *ENAS \citep{pham2018efficient}&19.43&4.6&0.5\\
        *AmoebaNet \citep{real2019regularized}&18.93&3.1&3150\\
    \hline
               ${}^{\dag}$DARTS-1st \citep{liu2018darts}  &20.52$\pm$0.31 &1.8 &0.4\\
    *GDAS \citep{DongY19}&18.38&3.4&0.2\\
    *R-DARTS \citep{ZelaESMBH20}&18.01$\pm$0.26&-&1.6
    \\
               *DARTS$^{-}$ \citep{abs-2009-01027}&17.51$\pm$0.25&3.3&0.4\\      ${}^{\dag}$DARTS$^{-}$ \citep{abs-2009-01027}& 18.97$\pm$0.16& 3.1&0.4\\
  ${}^{\Delta}$DARTS$^{+}$ \citep{abs-1909-06035}&17.11$\pm$0.43&3.8&0.2\\
      *DropNAS \citep{HongL0TWL020} & 16.39&4.4&0.7 \\
\hline
\hline
            *DARTS-2nd \citep{liu2018darts}  & 20.58$\pm$0.44&1.8&1.5 \\
            $\;\;$JL(DARTS2nd) & 18.92$\pm$0.17 & 2.4& 3.1 \\
             $\;\;$IL(DARTS2nd) (ours) & \textbf{17.12}$\pm$0.08 & 2.6& 3.2 \\
             \hline              *P-DARTS \citep{chen2019progressive}&17.49&3.6&0.3\\
             $\;\;$JL(PDARTS) &17.67$\pm$0.31 &3.5 &0.6\\
             $\;\;$IL(PDARTS) (ours)& \textbf{16.14}$\pm$0.17& 3.6&0.6\\
             \hline
                              $\dag$PC-DARTS \citep{abs-1907-05737} &17.96$\pm$0.15&3.9&0.1 \\
                        $\;\;$JL(PCDARTS) & 18.11$\pm$0.27& 3.9&0.2\\
                      $\;\;$IL(PCDARTS) (ours)&17.83$\pm$0.14 &3.8 &0.3\\
        \bottomrule
    \end{tabular}
    \label{tab:cifar100-il}
\end{table}

\begin{table}[t]
\caption{
    Classification error (\%) on the test set of CIFAR-10, number of parameters (millions) in the searched architecture, and search cost (GPU days).
     * denotes that the results are taken from DARTS$^{-}$ \citep{abs-2009-01027}, NoisyDARTS \citep{abs-2005-03566},  and DrNAS \citep{abs-2006-10355}.
   }
    \centering
    \begin{tabular}{l|ccc}
    \toprule
    Method& Error(\%)& Param(M) & Cost\\
    \midrule
    *DenseNet
    \citep{HuangLMW17}&3.46&25.6 &-\\
    \hline
     *HierEvol \citep{liu2017hierarchical}&3.75$\pm$0.12& 15.7 &300\\
    *NAONet-WS \citep{LuoTQCL18} & 3.53 & 3.1&0.4 \\
        *PNAS \citep{LiuZNSHLFYHM18} &3.41$\pm$0.09  &3.2& 225\\
        *ENAS \citep{pham2018efficient} &2.89 & 4.6  &0.5 \\
    *NASNet-A \citep{zoph2018learning} & 2.65 & 3.3& 1800\\
    *AmoebaNet-B \citep{real2019regularized} & 2.55$\pm$0.05 & 2.8&3150  \\
    \hline
                *DARTS-1st \citep{liu2018darts} &3.00$\pm$0.14&3.3&  0.4\\
        *R-DARTS \citep{ZelaESMBH20} &2.95$\pm$0.21  &- & 1.6 \\
            *GDAS \citep{DongY19}&2.93& 3.4& 0.2 \\
    *SNAS \citep{xie2018snas} &2.85 & 2.8& 1.5\\
    ${}^{\Delta}$DARTS$^{+}$ \citep{abs-1909-06035}&2.83$\pm$0.05&3.7&0.4\\
        *BayesNAS \citep{ZhouYWP19} &2.81$\pm$0.04 &3.4&0.2 \\
        *MergeNAS \citep{WangXYYHS20} &2.73$\pm$0.02 &2.9 & 0.2 \\
        *NoisyDARTS \citep{abs-2005-03566} &2.70$\pm$0.23&3.3  & 0.4 \\
            *ASAP \citep{NoyNRZDFGZ20} &2.68$\pm$0.11 & 2.5&0.2 \\
                *SDARTS
    \citep{abs-2002-05283}&2.61$\pm$0.02 & 3.3& 1.3 \\
         *DARTS$^{-}$ \citep{abs-2009-01027}&2.59$\pm$0.08&  3.5&0.4\\
         ${}^{\dag}$DARTS$^{-}$ \citep{abs-2009-01027}& 2.97$\pm$0.04& 3.3&0.4\\
            *DropNAS \citep{HongL0TWL020} &2.58$\pm$0.14 & 4.1&0.6 \\
    *FairDARTS \citep{abs-1911-12126} &2.54 &3.3 &0.4 \\
        *DrNAS \citep{abs-2006-10355} &2.54$\pm$0.03&4.0&  0.4\\
     \hline
        \hline
               *DARTS2nd \citep{liu2018darts} &2.76$\pm$0.09&3.3&  1.5\\ $\;\;$JL(DARTS2nd)   & 2.91$\pm$0.12  &2.4 & 3.1 \\
           $\;\;$IL(DARTS2nd) (ours)  & \textbf{2.62}$\pm$0.04 &2.6 & 3.2\\
           \hline
                    *PC-DARTS \citep{abs-1907-05737} &2.57$\pm$0.07&3.6& 0.1 \\
           $\;\;$JL(PCDARTS)  &2.63$\pm$0.05&3.9& 0.2\\
          $\;\;$IL(PCDARTS) (ours) &2.55$\pm$0.11&3.8&0.3 \\
          \hline 
                     *P-DARTS \citep{chen2019progressive} &2.50 &3.4&0.3\\ 
                      $\;\;$JL(PDARTS)  & 2.63$\pm$0.12 & 3.5& 0.6\\
                     $\;\;$IL(PDARTS) (ours) & 2.51$\pm$0.10 & 3.6& 0.6\\
        \bottomrule
    \end{tabular}
    \label{tab:c10-il}
\end{table}

\begin{table*}[t]
\caption{In interleaving learning experiments, top-1 and top-5 classification errors on the test set of ImageNet, number of weight parameters, and search cost. Results marked with * are taken from DARTS$^{-}$ \citep{abs-2009-01027} and DrNAS \citep{abs-2006-10355}. From top to bottom, in the first, second, and third block are human-designed networks, non-differentiable search methods, and differentiable search methods. 
    }
    \centering
    \begin{tabular}{l|cccc}
    \toprule
  \multirow{2}{*}{Method}   & Top-1  &Top-5 &Param & Cost \\
         & Error (\%) & Error (\%)&(M) & (GPU days)\\
    \midrule
    *Inception-v1 \citep{googlenet}&30.2 &10.1&6.6&- \\
    *MobileNet \citep{HowardZCKWWAA17} &  29.4& 10.5 &4.2&- \\
    *ShuffleNet 2$\times$ (v1) \citep{ZhangZLS18} &  26.4 &10.2 & 5.4&-\\
    *ShuffleNet 2$\times$ (v2) \citep{MaZZS18} &  25.1 &7.6 & 7.4&-\\
    \hline
    *NASNet-A \citep{zoph2018learning} &26.0 &8.4 &5.3 &1800 \\
    *PNAS \citep{LiuZNSHLFYHM18} &25.8 &8.1  &5.1 &225 \\
    *MnasNet-92 \citep{TanCPVSHL19} & 25.2 & 8.0& 4.4&1667\\
        *AmoebaNet-C \citep{real2019regularized} &  24.3 &7.6 &6.4&3150 \\
    \hline
     *SNAS \citep{xie2018snas} & 27.3 &9.2 &4.3 &1.5 \\
          *BayesNAS \citep{ZhouYWP19} &26.5 &8.9 &3.9&0.2 \\
                    *PARSEC \citep{abs-1902-05116} & 26.0 &8.4&5.6&1.0 \\
     *GDAS \citep{DongY19} &  26.0&8.5 &5.3 & 0.2\\
                 *DSNAS \citep{HuXZLSLL20} &25.7& 8.1 &- & -\\
          *SDARTS-ADV \citep{abs-2002-05283}&25.2& 7.8 &5.4& 1.3 \\
           *PC-DARTS \citep{abs-1907-05737} & 25.1 &7.8&5.3&0.1\\
                *ProxylessNAS \citep{cai2018proxylessnas} & 24.9 &7.5 &7.1 &8.3  \\
          *FairDARTS (CIFAR-10) \citep{abs-1911-12126} &24.9 &7.5 &4.8 &0.4 \\
     *FairDARTS (ImageNet) \citep{abs-1911-12126} &24.4 &7.4 &4.3 &3.0 \\
             *DrNAS \citep{abs-2006-10355} & 24.2 &7.3& 5.2&3.9\\
         *DARTS$^{+}$ (ImageNet) \citep{abs-1909-06035}& 23.9& 7.4&5.1&6.8\\
        *DARTS$^{-}$ \citep{abs-2009-01027}&23.8& 7.0&4.9&4.5\\
     *DARTS$^{+}$ (CIFAR-100) \citep{abs-1909-06035}&23.7& 7.2&5.1&0.2\\
     \hline
      \hline
         *DARTS2nd(CIFAR10) \citep{liu2018darts}  & 26.7 &8.7&4.7&1.5 \\
           $\;\;$JL(DARTS2nd,CIFAR10/100) & 26.4& 8.5  &3.5 & 3.1\\
        $\;\;$IL(DARTS2nd,CIFAR10/100) (ours) & \textbf{25.5} &\textbf{8.0} &3.8 & 3.2\\
        \hline
          *PDARTS(CIFAR10) \citep{chen2019progressive}&24.4 &7.4&4.9&0.3\\
             *PDARTS(CIFAR100) \citep{chen2019progressive}&24.7& 7.5&5.1&0.3\\
              $\;\;$JL(PDARTS,CIFAR10/100)  & 25.0 & 7.9 &5.1 &0.6 \\
           $\;\;$IL(PDARTS,CIFAR10/100) (ours) & \textbf{24.1} & \textbf{7.1} & 5.3&0.6 \\
         \bottomrule
    \end{tabular}
    \label{tab:imagenet-il}
\end{table*}

\subsubsection{Experimental Settings}
We follow the experimental protocol in \citep{liu2018darts}. Each experiment consists of a search phrase and an evaluation phrase. In the search phrase, an optimal architecture cell is searched. In the evaluation phrase, the searched cell is copied multiple times and these copies are stacked into a larger network. The larger network is trained from scratch. Each experiment was repeated 10 times with different random initialization.

In interleaving learning, we perform two tasks: image classification on CIFAR-100 and image classification on CIFAR-10, using two classification models $A$ and $B$. 
CIFAR-10 contains 10 classes and CIFAR-100 contains 100 classes. 
For CIFAR-10 and CIFAR-100, each of them is split into a 25K training set, a 25K validation set, and a 10K test set. The training and validation set of CIFAR-100 is used as $D_A^{(\textrm{tr})}$ and $D_A^{(\textrm{val})}$ respectively; the training and validation set of CIFAR-10 is used as $D_B^{(\textrm{tr})}$ and $D_B^{(\textrm{val})}$ respectively. For the architecture search space of the feature extractors, we experimented with the search spaces of DARTS~\citep{liu2018darts}, P-DARTS~\citep{chen2019progressive}, and PC-DARTS~\citep{abs-1907-05737}. These search spaces are composed of $3\times 3$ and $5\times 5$ (dilated) separable convolutions, $3\times 3$ max pooling, $3\times 3$ average pooling, zero, and  identity. 
For the CIFAR-100 classification head, we set it to a 100-way linear classifier. For the CIFAR-10 classification head, we set it to a 10-way linear classifier.

During architecture search, we perform two rounds of learning, with an order of CIFAR-100, CIFAR-10, CIFAR-100, CIFAR-10. The tradeoff parameter $\beta$ in interleaving learning was set to 100. The network of the feature extractor is a stack of 8 cells. Each cell has 7 nodes. We set the initial channel number to 16. We optimized the architecture variables using Adam~\citep{adam}. The learning rate was set to 3e-5 for IL-DARTS, 6e-4 for IL-P-DARTS, and 3e-3 for IL-PC-DARTS. The weight decay was set to 1e-3. 
We optimized the network weights using SGD. The initial learning rate was set to 0.025 for IL-DARTS and IL-P-DARTS and 0.1 for IL-PC-DARTS. The batch size was set to 64 for IL-DARTS and IL-P-DARTS and 256 for IL-PC-DARTS. The epoch number was set to 50 for IL-DARTS and IL-PC-DARTS and 25 for IL-P-DARTS. Weight decay was set to 3e-4 and momentum was set to  0.9. Cosine decay scheduler was used for scheduling the learning rate. The search was performed on a single Teslav100 GPU.

During architecture evaluation, the searched cell in interleaving learning is evaluated on CIFAR-10 and CIFAR-100 independently. For either dataset, 20 copies of the searched cell are composed into a larger network as the feature extractor, which is trained on the combination of training and validation sets and tested on the test set. The initial channel number was set to 36. We trained the network for 600 epochs, with a mini-batch size of 96 for IL-DARTS and 64 for IL-P-DARTS and IL-PC-DARTS. The evaluation on CIFAR-10 and CIFAR-100 was performed on a single Tesla v100 GPU. Given the architecture searched on CIFAR10/100, we also evaluate it on ImageNet. Specifically, 14 copies of the  searched  cell are composed into a larger network as the feature extractor, which is trained on 1.2M training images in ImageNet and tested on 50K testing images. We set the initial channel number to 48. The number of epochs was set to 250. The batch size was set to 1024. The training on ImageNet was performed on Tesla v100 GPUs.

\subsubsection{Experimental Results}
Table~\ref{tab:cifar100-il} and Table~\ref{tab:c10-il} shows the results on CIFAR-100 and CIFAR-10 respectively, including classification errors on the test set, number of model parameters, and search cost (GPU days). As can be seen, applying interleaving learning (IL) to DARTS-2nd, P-DARTS, and PC-DARTS significantly reduces the errors of these baselines approaches. For example, on CIFAR-100, applying IL to DARTS-2nd reduces the error from 20.58\% to 17.12\% and applying IL to P-DARTS reduces the error from 17.49\% to 16.14\%. As another example, on CIFAR-10, applying IL to DARTS-2nd reduces the error from 2.76\% to 2.62\%. These results show that interleaving learning can help to search better architectures. With interleaving learning, the search task on CIFAR-100 and CIAFR-10 can mutually benefit each other. The feature extractor $W_A$ trained on CIFAR-100 is used to initialize the feature extractor $W_B$ for CIFAR-10. Since $W_A$ is trained on CIFAR-100, it is better than random weights. Therefore, using $W_A$ to initialize $W_B$ is better than random initialization. Likewise, the $W_B$ trained on CIFAR-10 is used to initialize $W_A$ in the next round of training, which is better than random initialization. These two feature extractors mutually help each other to improve in the interleaving process. In baseline approaches, such a mechanism is missing. Therefore, interleaving learning achieves better performance than the baselines. 

One may wonder whether the performance gain of interleaving learning is due to more data (CIFAR100+CIFAR10) is used. To investigate this, we compare interleaving learning with a joint learning (JL) baseline with the following formulation:
\begin{equation}
    \begin{array}{ll}
       \textrm{min}_{T}  &  L(T, W^*_A(T), H^*_A(T), D_A^{(\textrm{val})})+  L(T, W^*_B(T),H^*_B(T), D_B^{(\textrm{val})})\\
        s.t. & W^*_A(T), H^*_A(T),W^*_B(T),H^*_B(T) =\textrm{argmin}_{W_A, H_A, W_B, H_B} \;\;L(T, W_A, H_A, D_A^{(\textrm{tr})})+\\
        &\qquad\qquad\qquad\qquad\qquad\qquad\qquad\qquad\qquad\qquad\qquad\qquad L(T, W_B, H_B, D_B^{(\textrm{tr})})
    \end{array}
\end{equation}
where $T$ denotes the architecture of the feature extractors; $W_A$ and $H_A$ denote the feature extractor weights and classification head of model $A$ (for CIFAR-100); $W_B$ and $H_B$ denote the feature extractor weights and classification head of model $B$ (for CIFAR-10). In the inner optimization problem (on the constraint), we train $W_A$, $H_A$, $W_B$, and $H_B$ by minimizing the training losses defined of both datasets, with the architecture fixed. In the outer optimization problem, we update the architecture by minimizing the validation losses of both datasets. In this formulation, the architecture is learned using both datasets. Table~\ref{tab:cifar100-il} and Table~\ref{tab:c10-il} show the results of this joint learning (JL) formulation. As can be seen, our IL method outperforms JL. For example, on CIFAR-100, when applied to DARTS-2nd, the error of IL is 17.12\% while that of JL is 18.92\%; when applied to P-DARTS, the error of IL is 16.14\% while that of JL is 17.67\%. As another example, on CIFAR-10, when applied to DARTS-2nd, the error of IL is 2.62\% while that of JL is 2.91\%. These results show that the performance gain of interleaving learning comes from the interleaving mechanism rather than due to more data is used. In the JL formulation, the weights of feature extractors of model $A$ and $B$ are trained independently. In contrast, in IL, the weights of feature extractors of model $A$ and $B$ mutually help each other to improve via pretraining in the interleaving process. Therefore, IL achieves better performance than JL. It is worth noting that while IL achieves better performance than baselines, it does not substantially increase parameter number or search cost.

Table~\ref{tab:imagenet-il} shows the results on ImageNet, including top-1 and top-5 errors, number of model parameters, and search cost (GPU days). From this table, we make the following observations. First, applying IL to DARTS and P-DARTS reduces the errors of these two baselines. For example, applying IL to DARTS-2nd reduces the top-1 error from 26.7\% to 25.5\% and reduces the top-5 error from 8.7\% to 8.0\%. This further demonstrates the effectiveness of IL in searching better architectures by encouraging two models to mutually help each other in the interleaving process. Second, IL achieves better performance than JL. For example, applied to DARTS-2nd, the top-1 and top-5 errors of IL are lower than those in JL. This further shows that the performance gain of IL comes from the interleaving process rather than leveraging more data. Third, while achieving better performance, IL does not substantially increase parameter number or search cost.

\subsection{Summary}
In this section, we apply Skillearn to formalize the interleaving learning (IL) skill of humans. In IL, a set of  models collaboratively learn a data encoder in an interleaving fashion: the encoder is trained by model 1 for a while, then passed to model 2 for further training, then model 3, and so on; after trained by all models, the encoder returns back to model 1 and is trained again, then moving to model 2, 3, etc. This process repeats for multiple rounds. Via interleaving, different models transfer their learned knowledge to each other to better represent data and avoid being stuck in bad local optimums.    
Experiments of neural architecture search  on  CIFAR-100 and CIFAR-10  demonstrate the effectiveness of interleaving learning.

\subsection{Related Works}
\subsubsection{Neural Architecture Search}
Neural architecture search (NAS) has achieved remarkable progress recently, which aims at searching for the optimal architecture of neural networks to achieve the best predictive performance.  In general, there are three paradigms of methods in NAS: reinforcement learning (RL) approaches~\citep{zoph2016neural,pham2018efficient,zoph2018learning}, evolutionary learning approaches~\citep{liu2017hierarchical,real2019regularized}, and differentiable  approaches~\citep{cai2018proxylessnas,liu2018darts,xie2018snas}. In RL-based approaches, a policy is learned to iteratively generate new architectures by maximizing a reward which is the accuracy on the validation set. Evolutionary learning approaches represent the architectures as individuals in a population. Individuals with high fitness scores (validation accuracy) have the privilege to generate offspring, which replaces individuals with low fitness scores. Differentiable  approaches adopt a network pruning strategy. On top of an over-parameterized network, the weights of connections between nodes are learned using gradient descent. Then weights close to zero are pruned later on. There have been many efforts devoted to improving differentiable NAS methods. 
 In P-DARTS \citep{chen2019progressive}, the depth of searched architectures is allowed to grow progressively during the training process.   Search space approximation and
regularization approaches are developed to reduce computational overheads and improve search stability.  PC-DARTS \citep{abs-1907-05737} reduces the redundancy in exploring the search space by sampling a small portion  of a super network. Operation search is performed in a subset of channels with the held-out part bypassed in a shortcut. Our proposed LCT framework can be applied to any differentiable NAS methods.

\subsubsection{Adversarial Learning}
Our proposed LPT involves a min-max optimization problem, which is analogous to that in adversarial learning. 
Adversarial learning~\citep{goodfellow2014generative} has been widely applied to 1) data generation~\citep{goodfellow2014generative,yu2017seqgan} where a discriminator tries to distinguish between  generated images and real images and a generator is trained to generate realistic data by making such a discrimination difficult to achieve; 2) domain adaptation~\citep{ganin2015unsupervised} where a discriminator tries to differentiate between source images and target images while the feature learner learns representations which make such a discrimination unachievable; 3) adversarial attack and defence~\citep{goodfellow2014explaining} where an attacker  adds small perturbations to the input data to alter the prediction outcome and the defender trains the model in a way that the prediction outcome remains the same given perturbed inputs. Different from these existing works, in our work, a tester aims to  create harder tests to ``fail" the learner while the learner learns to ``pass" however hard tests created by the tester. \citet{shu2020identifying} proposed to use an adversarial examiner to identify the weakness of a trained model. Our work differs from this one in that we  progressively re-train a learner model based on how it performs on the tests dynamically created by a tester model while the learner model in \citep{shu2020identifying} is fixed and not affected by the examination results.

\section{Conclusions}
In this paper, we develop a general framework called Skillearn to formalize humans' learning skills into machine-executable learning skills and leverage them to train better machine learning models. Our framework can flexibly formulate many learning skills of humans, by mapping from learners, learnable parameters, interaction functions,  learning stages, etc. in Skillearn to their counterparts in human learning. The formulated machine-executable learning skills can be applied to improve any ML model. In two case studies, we apply Skillearn to formalize two learning skills of humans -- learning by passing tests (LPT) and interleaving learning (IL). In LPT, a tester model dynamically creates tests with increasing levels of difficulty to evaluate a testee model; the testee continuously improves its architecture by passing however difficult tests created by the tester. In IL, a set of  models collaboratively learn a data encoder in an interleaving fashion: the encoder is trained by model 1 for a while, then passed to model 2 for further training, then model 3, and so on; after trained by all models, the encoder returns back to model 1 and is trained again, then moving to model 2, 3, etc. This process repeats for multiple rounds. Experiments on various datasets demonstrate that ML models trained by these two learning skills achieve significantly better performance. 

\bibliography{release}

\begin{thebibliography}{44}
\providecommand{\natexlab}[1]{#1}
\providecommand{\url}[1]{\texttt{#1}}
\expandafter\ifx\csname urlstyle\endcsname\relax
  \providecommand{\doi}[1]{doi: #1}\else
  \providecommand{\doi}{doi: \begingroup \urlstyle{rm}\Url}\fi

\bibitem[Cai et~al.(2019)Cai, Zhu, and Han]{cai2018proxylessnas}
Han Cai, Ligeng Zhu, and Song Han.
\newblock Proxylessnas: Direct neural architecture search on target task and
  hardware.
\newblock In \emph{{ICLR}}, 2019.

\bibitem[Casale et~al.(2019)Casale, Gordon, and Fusi]{abs-1902-05116}
Francesco~Paolo Casale, Jonathan Gordon, and Nicol{\'{o}} Fusi.
\newblock Probabilistic neural architecture search.
\newblock \emph{CoRR}, abs/1902.05116, 2019.

\bibitem[Chen and Hsieh(2020)]{abs-2002-05283}
Xiangning Chen and Cho{-}Jui Hsieh.
\newblock Stabilizing differentiable architecture search via perturbation-based
  regularization.
\newblock \emph{CoRR}, abs/2002.05283, 2020.

\bibitem[Chen et~al.(2020)Chen, Wang, Cheng, Tang, and Hsieh]{abs-2006-10355}
Xiangning Chen, Ruochen Wang, Minhao Cheng, Xiaocheng Tang, and Cho{-}Jui
  Hsieh.
\newblock Drnas: Dirichlet neural architecture search.
\newblock \emph{CoRR}, abs/2006.10355, 2020.

\bibitem[Chen et~al.(2019)Chen, Xie, Wu, and Tian]{chen2019progressive}
Xin Chen, Lingxi Xie, Jun Wu, and Qi~Tian.
\newblock Progressive differentiable architecture search: Bridging the depth
  gap between search and evaluation.
\newblock In \emph{ICCV}, 2019.

\bibitem[Chu et~al.(2019)Chu, Zhou, Zhang, and Li]{abs-1911-12126}
Xiangxiang Chu, Tianbao Zhou, Bo~Zhang, and Jixiang Li.
\newblock Fair {DARTS:} eliminating unfair advantages in differentiable
  architecture search.
\newblock \emph{CoRR}, abs/1911.12126, 2019.

\bibitem[Chu et~al.(2020{\natexlab{a}})Chu, Wang, Zhang, Lu, Wei, and
  Yan]{abs-2009-01027}
Xiangxiang Chu, Xiaoxing Wang, Bo~Zhang, Shun Lu, Xiaolin Wei, and Junchi Yan.
\newblock {DARTS-:} robustly stepping out of performance collapse without
  indicators.
\newblock \emph{CoRR}, abs/2009.01027, 2020{\natexlab{a}}.

\bibitem[Chu et~al.(2020{\natexlab{b}})Chu, Zhang, and Li]{abs-2005-03566}
Xiangxiang Chu, Bo~Zhang, and Xudong Li.
\newblock Noisy differentiable architecture search.
\newblock \emph{CoRR}, abs/2005.03566, 2020{\natexlab{b}}.

\bibitem[Deng et~al.(2009)Deng, Dong, Socher, Li, Li, and
  Fei-Fei]{deng2009imagenet}
Jia Deng, Wei Dong, Richard Socher, Li-Jia Li, Kai Li, and Li~Fei-Fei.
\newblock Imagenet: A large-scale hierarchical image database.
\newblock In \emph{2009 IEEE conference on computer vision and pattern
  recognition}, pages 248--255. Ieee, 2009.

\bibitem[Dong and Yang(2019)]{DongY19}
Xuanyi Dong and Yi~Yang.
\newblock Searching for a robust neural architecture in four {GPU} hours.
\newblock In \emph{{CVPR}}, 2019.

\bibitem[Ganin and Lempitsky(2015)]{ganin2015unsupervised}
Yaroslav Ganin and Victor Lempitsky.
\newblock Unsupervised domain adaptation by backpropagation.
\newblock In \emph{International Conference on Machine Learning}, pages
  1180--1189, 2015.

\bibitem[Goodfellow et~al.(2014{\natexlab{a}})Goodfellow, Pouget-Abadie, Mirza,
  Xu, Warde-Farley, Ozair, Courville, and Bengio]{goodfellow2014generative}
Ian Goodfellow, Jean Pouget-Abadie, Mehdi Mirza, Bing Xu, David Warde-Farley,
  Sherjil Ozair, Aaron Courville, and Yoshua Bengio.
\newblock Generative adversarial nets.
\newblock In \emph{Advances in neural information processing systems}, pages
  2672--2680, 2014{\natexlab{a}}.

\bibitem[Goodfellow et~al.(2014{\natexlab{b}})Goodfellow, Shlens, and
  Szegedy]{goodfellow2014explaining}
Ian~J Goodfellow, Jonathon Shlens, and Christian Szegedy.
\newblock Explaining and harnessing adversarial examples.
\newblock \emph{arXiv preprint arXiv:1412.6572}, 2014{\natexlab{b}}.

\bibitem[He et~al.(2016{\natexlab{a}})He, Zhang, Ren, and Sun]{he2016deep}
Kaiming He, Xiangyu Zhang, Shaoqing Ren, and Jian Sun.
\newblock Deep residual learning for image recognition.
\newblock In \emph{CVPR}, 2016{\natexlab{a}}.

\bibitem[He et~al.(2016{\natexlab{b}})He, Zhang, Ren, and Sun]{resnet}
Kaiming He, Xiangyu Zhang, Shaoqing Ren, and Jian Sun.
\newblock Deep residual learning for image recognition.
\newblock In \emph{CVPR}, 2016{\natexlab{b}}.

\bibitem[He et~al.(2019)He, Fan, Wu, Xie, and Girshick]{he2019moco}
Kaiming He, Haoqi Fan, Yuxin Wu, Saining Xie, and Ross Girshick.
\newblock Momentum contrast for unsupervised visual representation learning.
\newblock \emph{arXiv preprint arXiv:1911.05722}, 2019.

\bibitem[Hong et~al.(2020)Hong, Li, Zhang, Tang, Wang, Li, and
  Yu]{HongL0TWL020}
Weijun Hong, Guilin Li, Weinan Zhang, Ruiming Tang, Yunhe Wang, Zhenguo Li, and
  Yong Yu.
\newblock Dropnas: Grouped operation dropout for differentiable architecture
  search.
\newblock In \emph{{IJCAI}}, 2020.

\bibitem[Howard et~al.(2017)Howard, Zhu, Chen, Kalenichenko, Wang, Weyand,
  Andreetto, and Adam]{HowardZCKWWAA17}
Andrew~G. Howard, Menglong Zhu, Bo~Chen, Dmitry Kalenichenko, Weijun Wang,
  Tobias Weyand, Marco Andreetto, and Hartwig Adam.
\newblock Mobilenets: Efficient convolutional neural networks for mobile vision
  applications.
\newblock \emph{CoRR}, abs/1704.04861, 2017.

\bibitem[Hu et~al.(2020)Hu, Xie, Zheng, Liu, Shi, Liu, and Lin]{HuXZLSLL20}
Shoukang Hu, Sirui Xie, Hehui Zheng, Chunxiao Liu, Jianping Shi, Xunying Liu,
  and Dahua Lin.
\newblock {DSNAS:} direct neural architecture search without parameter
  retraining.
\newblock In \emph{{CVPR}}, 2020.

\bibitem[Huang et~al.(2017)Huang, Liu, van~der Maaten, and
  Weinberger]{HuangLMW17}
Gao Huang, Zhuang Liu, Laurens van~der Maaten, and Kilian~Q. Weinberger.
\newblock Densely connected convolutional networks.
\newblock In \emph{{CVPR}}, 2017.

\bibitem[Kingma and Ba(2014)]{adam}
Diederik Kingma and Jimmy Ba.
\newblock Adam: A method for stochastic optimization.
\newblock \emph{International Conference on Learning Representations}, 12 2014.

\bibitem[Liang et~al.(2019{\natexlab{a}})Liang, Zhang, Sun, He, Huang, Zhuang,
  and Li]{abs-1909-06035}
Hanwen Liang, Shifeng Zhang, Jiacheng Sun, Xingqiu He, Weiran Huang, Kechen
  Zhuang, and Zhenguo Li.
\newblock {DARTS+:} improved differentiable architecture search with early
  stopping.
\newblock \emph{CoRR}, abs/1909.06035, 2019{\natexlab{a}}.

\bibitem[Liang et~al.(2019{\natexlab{b}})Liang, Zhang, Sun, He, Huang, Zhuang,
  and Li]{liang2019darts+}
Hanwen Liang, Shifeng Zhang, Jiacheng Sun, Xingqiu He, Weiran Huang, Kechen
  Zhuang, and Zhenguo Li.
\newblock Darts+: Improved differentiable architecture search with early
  stopping.
\newblock \emph{arXiv preprint arXiv:1909.06035}, 2019{\natexlab{b}}.

\bibitem[Lin et~al.(2014)Lin, Maire, Belongie, Hays, Perona, Ramanan,
  Doll{\'a}r, and Zitnick]{coco}
Tsung-Yi Lin, Michael Maire, Serge Belongie, James Hays, Pietro Perona, Deva
  Ramanan, Piotr Doll{\'a}r, and C~Lawrence Zitnick.
\newblock Microsoft coco: Common objects in context.
\newblock In \emph{ECCV}, 2014.

\bibitem[Liu et~al.(2018{\natexlab{a}})Liu, Zoph, Neumann, Shlens, Hua, Li,
  Fei{-}Fei, Yuille, Huang, and Murphy]{LiuZNSHLFYHM18}
Chenxi Liu, Barret Zoph, Maxim Neumann, Jonathon Shlens, Wei Hua, Li{-}Jia Li,
  Li~Fei{-}Fei, Alan~L. Yuille, Jonathan Huang, and Kevin Murphy.
\newblock Progressive neural architecture search.
\newblock In \emph{{ECCV}}, 2018{\natexlab{a}}.

\bibitem[Liu et~al.(2018{\natexlab{b}})Liu, Simonyan, Vinyals, Fernando, and
  Kavukcuoglu]{liu2017hierarchical}
Hanxiao Liu, Karen Simonyan, Oriol Vinyals, Chrisantha Fernando, and Koray
  Kavukcuoglu.
\newblock Hierarchical representations for efficient architecture search.
\newblock In \emph{{ICLR}}, 2018{\natexlab{b}}.

\bibitem[Liu et~al.(2019)Liu, Simonyan, and Yang]{liu2018darts}
Hanxiao Liu, Karen Simonyan, and Yiming Yang.
\newblock {DARTS:} differentiable architecture search.
\newblock In \emph{{ICLR}}, 2019.

\bibitem[Luo et~al.(2018)Luo, Tian, Qin, Chen, and Liu]{LuoTQCL18}
Renqian Luo, Fei Tian, Tao Qin, Enhong Chen, and Tie{-}Yan Liu.
\newblock Neural architecture optimization.
\newblock In \emph{NeurIPS}, 2018.

\bibitem[Ma et~al.(2018)Ma, Zhang, Zheng, and Sun]{MaZZS18}
Ningning Ma, Xiangyu Zhang, Hai{-}Tao Zheng, and Jian Sun.
\newblock Shufflenet {V2:} practical guidelines for efficient {CNN}
  architecture design.
\newblock In \emph{{ECCV}}, 2018.

\bibitem[Noy et~al.(2020)Noy, Nayman, Ridnik, Zamir, Doveh, Friedman, Giryes,
  and Zelnik]{NoyNRZDFGZ20}
Asaf Noy, Niv Nayman, Tal Ridnik, Nadav Zamir, Sivan Doveh, Itamar Friedman,
  Raja Giryes, and Lihi Zelnik.
\newblock {ASAP:} architecture search, anneal and prune.
\newblock In \emph{{AISTATS}}, 2020.

\bibitem[Pham et~al.(2018)Pham, Guan, Zoph, Le, and Dean]{pham2018efficient}
Hieu Pham, Melody~Y. Guan, Barret Zoph, Quoc~V. Le, and Jeff Dean.
\newblock Efficient neural architecture search via parameter sharing.
\newblock In \emph{{ICML}}, 2018.

\bibitem[Real et~al.(2019)Real, Aggarwal, Huang, and Le]{real2019regularized}
Esteban Real, Alok Aggarwal, Yanping Huang, and Quoc~V Le.
\newblock Regularized evolution for image classifier architecture search.
\newblock In \emph{Proceedings of the aaai conference on artificial
  intelligence}, volume~33, pages 4780--4789, 2019.

\bibitem[Shu et~al.(2020)Shu, Liu, Qiu, and Yuille]{shu2020identifying}
Michelle Shu, Chenxi Liu, Weichao Qiu, and Alan Yuille.
\newblock Identifying model weakness with adversarial examiner.
\newblock In \emph{Proceedings of the AAAI Conference on Artificial
  Intelligence}, volume~34, pages 11998--12006, 2020.

\bibitem[Szegedy et~al.(2015)Szegedy, Liu, Jia, Sermanet, Reed, Anguelov,
  Erhan, Vanhoucke, and Rabinovich]{googlenet}
Christian Szegedy, Wei Liu, Yangqing Jia, Pierre Sermanet, Scott Reed, Dragomir
  Anguelov, Dumitru Erhan, Vincent Vanhoucke, and Andrew Rabinovich.
\newblock Going deeper with convolutions.
\newblock In \emph{CVPR}, 2015.

\bibitem[Tan et~al.(2019)Tan, Chen, Pang, Vasudevan, Sandler, Howard, and
  Le]{TanCPVSHL19}
Mingxing Tan, Bo~Chen, Ruoming Pang, Vijay Vasudevan, Mark Sandler, Andrew
  Howard, and Quoc~V. Le.
\newblock Mnasnet: Platform-aware neural architecture search for mobile.
\newblock In \emph{{CVPR}}, 2019.

\bibitem[Wang et~al.(2020)Wang, Xue, Yan, Yang, Hu, and Sun]{WangXYYHS20}
Xiaoxing Wang, Chao Xue, Junchi Yan, Xiaokang Yang, Yonggang Hu, and Kewei Sun.
\newblock Mergenas: Merge operations into one for differentiable architecture
  search.
\newblock In \emph{{IJCAI}}, 2020.

\bibitem[Xie et~al.(2019)Xie, Zheng, Liu, and Lin]{xie2018snas}
Sirui Xie, Hehui Zheng, Chunxiao Liu, and Liang Lin.
\newblock {SNAS:} stochastic neural architecture search.
\newblock In \emph{{ICLR}}, 2019.

\bibitem[Xu et~al.(2020)Xu, Xie, Zhang, Chen, Qi, Tian, and
  Xiong]{abs-1907-05737}
Yuhui Xu, Lingxi Xie, Xiaopeng Zhang, Xin Chen, Guo{-}Jun Qi, Qi~Tian, and
  Hongkai Xiong.
\newblock {PC-DARTS:} partial channel connections for memory-efficient
  architecture search.
\newblock In \emph{{ICLR}}, 2020.

\bibitem[Yu et~al.(2017)Yu, Zhang, Wang, and Yu]{yu2017seqgan}
Lantao Yu, Weinan Zhang, Jun Wang, and Yong Yu.
\newblock Seqgan: Sequence generative adversarial nets with policy gradient.
\newblock In \emph{AAAI}, 2017.

\bibitem[Zela et~al.(2020)Zela, Elsken, Saikia, Marrakchi, Brox, and
  Hutter]{ZelaESMBH20}
Arber Zela, Thomas Elsken, Tonmoy Saikia, Yassine Marrakchi, Thomas Brox, and
  Frank Hutter.
\newblock Understanding and robustifying differentiable architecture search.
\newblock In \emph{{ICLR}}, 2020.

\bibitem[Zhang et~al.(2018)Zhang, Zhou, Lin, and Sun]{ZhangZLS18}
Xiangyu Zhang, Xinyu Zhou, Mengxiao Lin, and Jian Sun.
\newblock Shufflenet: An extremely efficient convolutional neural network for
  mobile devices.
\newblock In \emph{{CVPR}}, 2018.

\bibitem[Zhou et~al.(2019)Zhou, Yang, Wang, and Pan]{ZhouYWP19}
Hongpeng Zhou, Minghao Yang, Jun Wang, and Wei Pan.
\newblock Bayesnas: {A} bayesian approach for neural architecture search.
\newblock In \emph{{ICML}}, 2019.

\bibitem[Zoph and Le(2017)]{zoph2016neural}
Barret Zoph and Quoc~V. Le.
\newblock Neural architecture search with reinforcement learning.
\newblock In \emph{{ICLR}}, 2017.

\bibitem[Zoph et~al.(2018)Zoph, Vasudevan, Shlens, and Le]{zoph2018learning}
Barret Zoph, Vijay Vasudevan, Jonathon Shlens, and Quoc~V Le.
\newblock Learning transferable architectures for scalable image recognition.
\newblock In \emph{CVPR}, 2018.

\end{thebibliography}

\end{document}